\documentclass[10pt, compsocconf]{formatting/IEEEtran}
\usepackage{mathptmx}
\newcommand{\iscasubmissionnumber}{129}

\usepackage{formatting/hadianeh}
\usepackage{formatting/macros}
\usepackage{formatting/defs}
\usepackage{formatting/numbers}
\newcommand{\ignore}[1]{}
\usepackage{fancyhdr}
\usepackage[normalem]{ulem}
\usepackage{hyperref}
\usepackage{subfig}
\usepackage{tikz}
\usepackage[subtle]{savetrees}
\usepackage{flushend}

\let\OLDthebibliography\thebibliography
\renewcommand\thebibliography[1]{
  \OLDthebibliography{#1}
  \setlength{\parskip}{0pt}
  \setlength{\itemsep}{0pt plus 0.3ex}
}
\renewcommand{\IEEEbibitemsep}{0pt plus 0.5pt}
\makeatletter
\IEEEtriggercmd{\reset@font\normalfont\fontsize{8pt}{8.40pt}\selectfont}
\makeatother
\IEEEtriggeratref{1}

\begin{document}\sloppy

\title{\vspace{-1ex}\bitfusiontitle: Bit-Level Dynamically Composable Architecture for Accelerating Deep Neural Networks}

\author{
\vspace{-1ex}
\begin{tabular}{cccc}
Hardik Sharma\IEEEauthorrefmark{1}\quad\quad\quad & Jongse Park\IEEEauthorrefmark{1}\quad\quad\quad & Naveen Suda\IEEEauthorrefmark{2}\quad\quad\quad & Liangzhen Lai\IEEEauthorrefmark{2}
\\
\end{tabular}
\vspace{-1ex}
\\
\begin{tabular}{ccc}
Benson Chau\IEEEauthorrefmark{1}\quad\quad\quad & Vikas Chandra\IEEEauthorrefmark{2}\quad\quad\quad & Hadi Esmaeilzadeh\IEEEauthorrefmark{3}
\\
\vspace{-2ex}
\\
\multicolumn{3}{c}{\fontsize{12}{16}\selectfont{}\textbf{A}lternative \textbf{C}omputing \textbf{T}echnologies ({\color[HTML]{0B6121}{\textbf{ACT}}}) Lab}
\end{tabular}
\IEEEauthorblockN{}
\begin{tabular}{ccc}
\fontsize{12}{16}\selectfont{}\IEEEauthorrefmark{1}Georgia Institute of Technology\quad\quad\quad &\fontsize{12}{16}\selectfont{}\IEEEauthorrefmark{2}Arm, Inc.\quad\quad\quad &\fontsize{12}{16}\selectfont{}\IEEEauthorrefmark{3}University of California, San Diego
\end{tabular}
\begin{tabular}{ccc}
\textcolor{blue}{\textsf{
\{\href{mailto:hsharma@gatech.edu}{hsharma},\href{mailto:jspark@gatech.edu}{jspark},\href{mailto:ben.chau@gatech.edu}{ben.chau}\}@gatech.edu}}\quad\quad
&\textcolor{blue}{\textsf{\{\href{mailto:naveen.suda@arm.com}{naveen.suda},\href{mailto:liangzhen.Lai@arm.com}{liangzhen.lai},\href{mailto:vikas.chandra@arm.com}{vikas.chandra}\}@arm.com}}\quad\quad
&\textcolor{blue}{\textsf{
\href{mailto:hadi@eng.ucsd.edu}{hadi@eng.ucsd.edu}}}
\end{tabular}
\vspace{-7ex}
}

\fancypagestyle{firstpage}{
\fancyhf{}   
\chead{\it Appears in the Proceedings of the 45$^{th}$ International Symposium on Computer Architecture (ISCA), 2018}
}

\maketitle
\thispagestyle{firstpage}
\begin{abstract}
Hardware acceleration of Deep Neural Networks (DNNs) aims to tame their enormous compute intensity.
Fully realizing the potential of acceleration in this domain requires understanding and leveraging algorithmic properties of DNNs.
This paper builds upon the algorithmic insight that bitwidth of operations in DNNs can be reduced without compromising their classification accuracy.
However, to prevent loss of accuracy, the bitwidth varies significantly across DNNs and it may even be adjusted for each layer individually.
Thus, a fixed-bitwidth accelerator would either offer limited benefits to accommodate the worst-case bitwidth requirements, or inevitably lead to a degradation in final accuracy.
To alleviate these deficiencies, this work introduces dynamic bit-level fusion/decomposition as a new dimension in the design of DNN accelerators.
We explore this dimension by designing \bitfusion, a bit-flexible accelerator, that constitutes an array of bit-level processing elements that dynamically fuse to match the bitwidth of individual DNN layers.
This flexibility in the architecture enables minimizing the computation and the communication at the finest granularity possible with no loss in accuracy.
We evaluate the benefits of \bitfusion using eight real-world feed-forward and recurrent DNNs.
The proposed microarchitecture is implemented in Verilog and synthesized in 45~nm technology.
Using the synthesis results and cycle accurate simulation, we compare the benefits of \bitfusion to two state-of-the-art DNN accelerators, Eyeriss~\cite{eyeriss:isca:2016} and Stripes~\cite{stripes:micro:2016}.
In the same area, frequency, and process technology, \bitfusion offers \eyerissPerfAvg speedup and \eyerissEnergyAvg energy savings over Eyeriss.
Compared to Stripes, \bitfusion provides \stripesPerfAvg speedup and \stripesEnergyAvg energy reduction at 45~nm node when \bitfusion area and frequency are set to those of Stripes.
Scaling to GPU technology node of 16~nm, \bitfusion almost matches the performance of a 250-Watt \titanxp, which uses $8$-bit vector instructions, while \bitfusion merely consumes 895 milliwatts of power.

\end{abstract}

\section{Introduction}
\label{sec:intro}

Advances in high-performance computer architecture design has been a major driver for the rapid evolution of Deep Neural Networks (DNN).
Due to their insatiable demand for compute power, naturally, both the research community~\cite{
eyeriss:isca:2016,
eyeriss:jssc:2017,
tetris:asplos:2017,
eie:isca:2016,
stripes:micro:2016,
tartan:arxiv:2017,
dadiannao:micro:2014,
diannao:asplos:2014,
pudiannao:asplos:2015,
shidiannao:isca:2015, 
neurocube:isca:2016,
minerva:isca:2016,
cnvlutin:isca:2016,
cambricon:isca:2016,
cambricon-x:micro:2016,
240gopsmobile:cvprw:2014,
iotcnn:isscc:2016,
riscintconv:date:2015,
deepburning:dac:2016,
cbrain:dac:2016,
dnnoptimizing:fpga:2015,
dnnweaver:micro:2016,
fusedlayercnn:micro:2016,
openclcnn:fpga:2016,
embedfpgacnn:fpga:2016,
isaac:isca:2016,
prime:isca:2016,
pipelayer:hpca:2017}
as well the industry~\cite{brainwave:hotchips:2017, tpu:isca:2017, apple-a11bionic:wiki:2017} have turned to accelerators to accommodate modern DNN computation.
However, the algorithmic properties of DNNs have not fully been utilized to push the envelope on their acceleration efficiency and performance.

To that end, we leverage the following three algorithmic properties of DNNs to introduce a novel acceleration architecture, called \bitfusion.
(1) DNNs are mostly a collection of massively parallel multiply-adds.
(2) The bitwidth of these operations can be reduced with no loss in accuracy~\cite{dorefa:arxiv:2016, zhu2016trained, li2016ternary, qnn:arxiv:2016, wrpn}.
(3) However, to preserve accuracy, the bitwidth varies significantly across DNNs and may even be adjusted for each layer individually.
Thus, a fixed-bitwidth accelerator design would either yield limited benefits to accommodate the worst-case bitwidth requirements, or inevitably lead to a degradation in final accuracy.
To alleviate these deficiencies, \bitfusion introduces the concept of runtime bit-level fusion/decomposition as a new dimension in the design of DNN accelerators. 
We explore this dimension by designing a bit-flexible accelerator, which comprises an array of processing engines that fuse at the bit-level granularity to match the bitwidth of the individual DNN layers.

\begin{figure}
	\centering
	\includegraphics[width=0.95\linewidth]{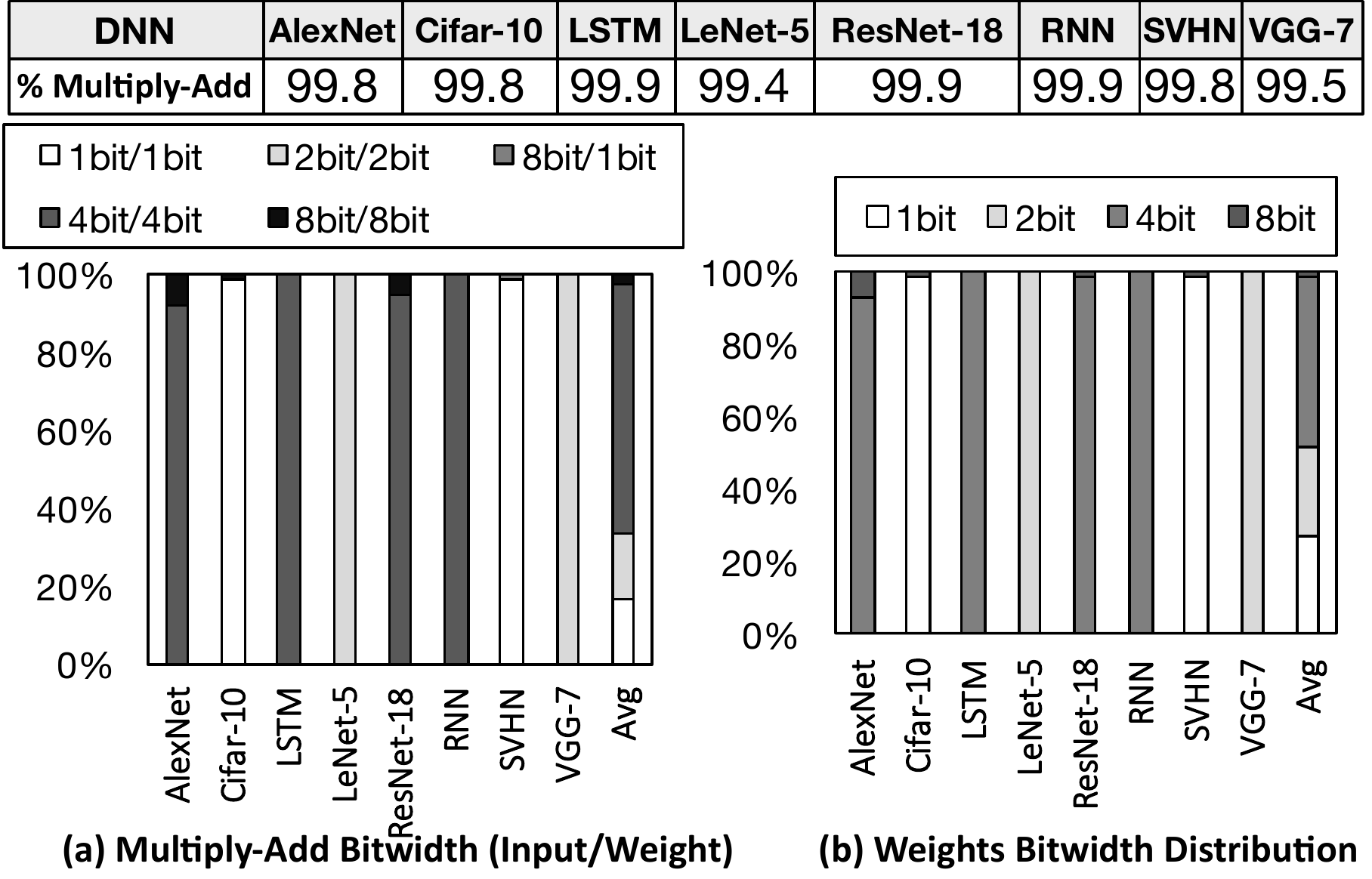}
	\vspace{-1ex}
	\caption{Bitwidth variation across real-world DNNs.}
	\vspace{-3.5ex}
	\label{fig:motivation}
\end{figure}

The bit-level flexibility in the architecture enables minimizing the computation and the communication at the finest granularity possible with no loss in accuracy.
As such, the following three insights both motivate and guide \bitfusion.

First, the number of bit-level operations required for the multiply operator is proportional to the product of the operands' bitwidths and scales linearly for the addition operator.
Therefore, matching the bitwidth of the multiply-add units to the reduced bitwidth of the DNN layers, almost quadratically reduces the bit-level computations.
This strategy will significantly affect the acceleration since the large majority of DNN operations ($>$ 99\%) are multiply-adds as shown in the table included in Figure~\ref{fig:motivation}.
For instance, each single image classification with AlexNet~\cite{wrpn} requires a total of 2682 million operations, of which 99.86\% (2678 million) are multiply-adds.
To this end, the compute units of \bitfusion can dynamically fuse or decompose to match the bitwidth of each individual multiply-add operand without requiring the operands to be encoded in the same bitwidth.

Second, energy consumption for DNN acceleration is usually dominated by data accesses to on-chip storage and  off-chip memory~\cite{tetris:asplos:2017, eyeriss:isca:2016, eyeriss:jssc:2017}.
Therefore, \bitfusion comes with encoding and memory access logic that stores and retrieves the values in the lowest required bitwidth.
This logic reduces the overall number of bits read or written to on-chip and off-chip memory, proportionally reducing the energy dissipation of memory accesses.
Furthermore, this strategy increases the effective on-chip storage capacity.

Third, \bitfusion builds upon the extensive prior work that shows DNNs can operate with reduced bitwidth without degradation in classification accuracy~\cite{dorefa:arxiv:2016, qnn:arxiv:2016, li2016ternary, zhu2016trained, envision:isscc:2017, stripes:micro:2016}.
This opportunity exists across different classes of real-world DNNs, 
as shown in Figure~\ref{fig:motivation}. 
One category is Convolutional Neural Networks (CNNs) that usually use convolution and pooling layers followed by a stack of fully-connected layers.
AlexNet, Cifar-10, LeNet-5, ResNet-18, SVHN, and VGG-7 in Figure~\ref{fig:motivation} belong to this category.
Recurrent Neural Networks (RNN) are another sub-class of DNNs that use recurrent layers including Long Short Term Memory (LSTM) and vanilla RNN layers to extract \emph{temporal} features from time-varying data.
The RNN and LSTM benchmark DNNs in Figure~\ref{fig:motivation} represent these categories.
Furthermore, as the table in Figure~\ref{fig:motivation} shows, most operations in DNNs ($>$ 99\%), regardless of their categories, are multiply-adds.
As Figure~\ref{fig:motivation}(a) illustrates, on average, 97.3\% of multiply-adds require four or fewer bits and even in some DNNs a large fraction of the operations can be done with bitwidth equal to one.
More interestingly, the bitwidths vary within and across DNNs to guarantee no loss of accuracy.
Such a variation is not limited to the intermediate operands and exists in trained weights as illustrated in Figure~\ref{fig:motivation}(b).
To exploit this property, a programmable accelerator needs to offer bit-level flexibility at runtime, which leads us to \bitfusion.

To harvest the aforementioned opportunities, this paper makes the following contributions and realizes a new dimension in the design of DNN accelerators.

\begin{enumpacked}
	\item \textbf{Dynamic bit-level fusion and decomposition.} 
	The paper introduces and explores the dimension of bit-level flexible DNN accelerator architectures, \bitfusion, that dynamically matches bit-level composable processing engines to the varying bitwidths required by DNN layers. 
	By offering this flexibility, \bitfusion aims to minimize the computation and communication required by a DNN at the bit granularity on a per layer basis.
	
	\item \textbf{Microarchitecture design for bit-level composability.} To explore \bitfusion, we design and implement a DNN accelerator using a novel bit-flexible computation unit, called \bricks.
	The accelerator supports both feed-forward (CNN) and recurrent (LSTM and RNN) layers.
	A 2D array of \bricks constructs a fusible processing engine that can  perform the DNN computation at various bitwidths.
	The microarchitecture also comes with a storage logic that allows feeding the \bricks with different bitwidth operands.
			
	\item \textbf{Hardware-software abstractions for bit-flexible acceleration.} To enable DNN applications to take advantage of these unique bit-level fusion capabilities, we propose a block-structured instruction set architecture, called \fusionisa. To amortize the cost of programmability, \fusionisa expresses operations of DNN layers as bit-flexible instruction blocks with iterative semantics.
\end{enumpacked}

These three contributions define the novel architecture of \bitfusion, a possible microarchitecture implementation, and the hardware-software abstractions to offer bit-level flexibility.
Other complementary and inspiring works have explored bit serial computation~\cite{stripes:micro:2016, tartan:arxiv:2017} without exploring the fusion dimension.
In contrast, \bitfusion \emph{spatially} fuses a group of \bricks together, to collectively execute operations at different bitwidths.
Using eight real-world feed-forward and recurrent real-world DNNs, we evaluate the benefits of \bitfusion.
We implemented the proposed microarchitecture in Verilog and synthesized in 45~nm technology.
Using the synthesis results and cycle accurate simulation, we compare the benefits of \bitfusion to two state-of-the-art DNN accelerators, Eyeriss~\cite{eyeriss:isca:2016} and Stripes~\cite{stripes:micro:2016}.
The latter is an optimized bit-serial architecture.
In the same area, frequency, and technology node, \bitfusion offers \eyerissPerfAvg speedup and \eyerissEnergyAvg energy savings over Eyeriss.
Compared to Stripes~\cite{stripes:micro:2016}, \bitfusion provides \stripesPerfAvg speedup and \stripesEnergyAvg energy reduction at 45~nm node when \bitfusion area and frequency are set to those of Stripes.
Scaling to GPU technology node of 16~nm, \bitfusion provides a \bitfusionPerfAvgOverTX speedup over the Jetson TX2 mobile GPU.
Further, \bitfusion almost matches the performance of a 250-Watt \titanxp, which uses $8$-bit vector instructions, while \bitfusion merely consumes 895 milliwatts of power.
\vspace{-1ex}
\section{\bitfusionsection Architecture}
\label{sec:arch}

\begin{figure}
	\centering
	\includegraphics[width=0.95\linewidth]{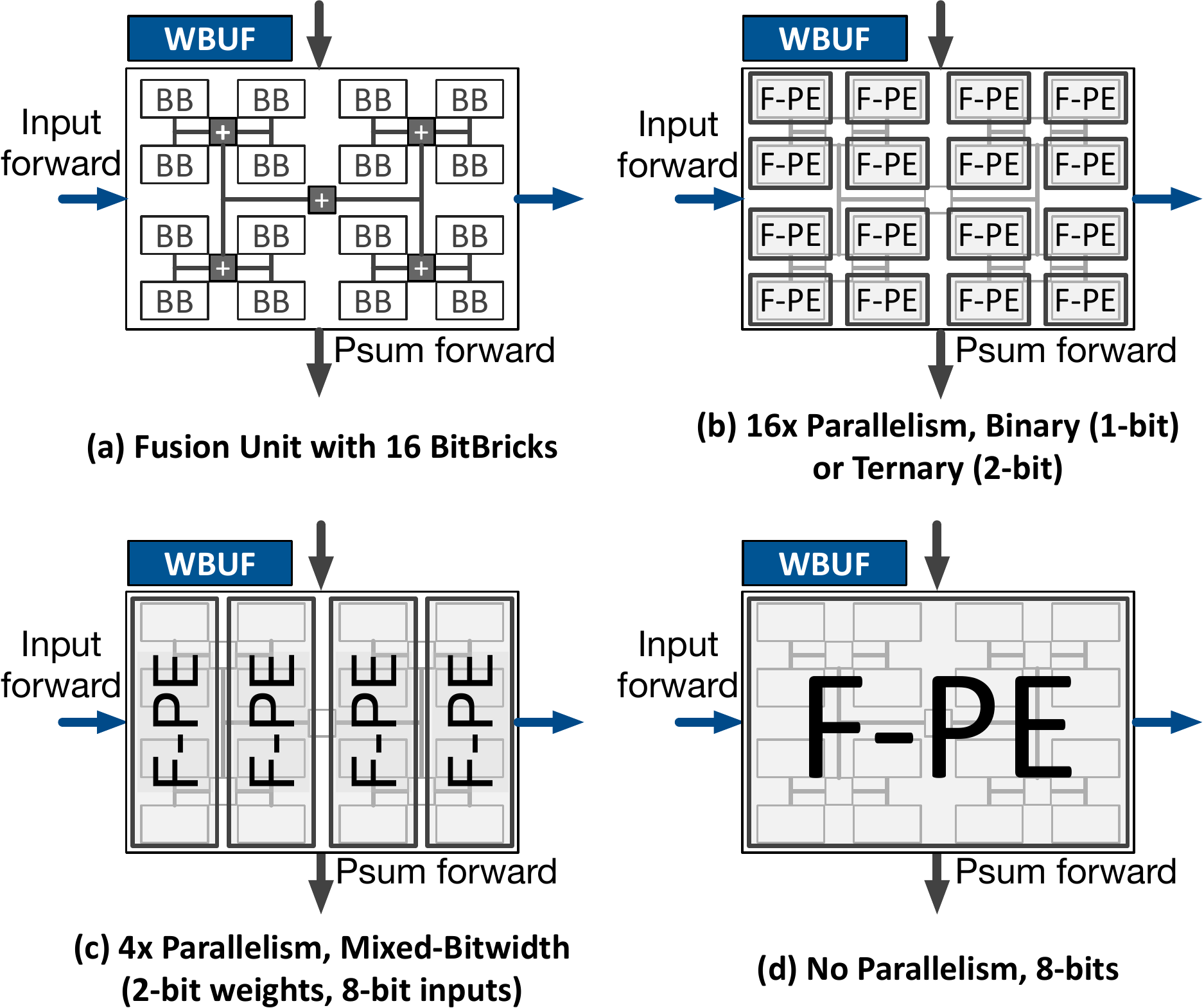}
	\vspace{-0.25ex}
	\caption{Dynamic composition of \bricks (BBs) in a \fusionunit to construct Fused Processing Engines (Fused-PE), shown as F-PE.}
	\label{fig:bitfusion}
	\vspace{-3ex}
\end{figure}

To minimize the computation and communication at the finest granularity, \bitfusion dynamically matches the architecture of the accelerator to the bitwidth required for the DNN, which may vary layer by layer, without any loss in accuracy.
As such, \bitfusion is a collection of  bit-level computational elements, called \bricks, that dynamically compose to \emph{logically} construct Fused Processing Engines (\fusedpe) that execute DNN operations with the required bitwidth.
Specifically, \fusedpes provide bit-level flexibility for multiply-adds, which are the dominant operations across all types of DNNs.
Below, we discuss how \bricks can be dynamically fused together to support a range of bitwidths, yet provide a significant increase in parallelism when operating at lower bitwidths.

\vspace{-1ex}
\subsection{Bit-Level Flexibility via Dynamic Fusion}
As depicted in Figure~\ref{fig:bitfusion}, \bitfusion arranges the \bitbricks in a 2-dimensional \emph{physical} grouping, called \fusionunit. 
Each \bitbrick in a \fusionunit can perform individual binary (0, +1) and ternary (-1, 0, +1) multiply-add operations.
As Figure~\ref{fig:bitfusion} shows, the \bricks \emph{logically} fuse together at run-time to form Fused Processing Engines (\fusedpes) that match the bitwidths required by the multiply-add operations of a DNN layer.
The \bricks in a \fusionunit multiply an incoming variable-bitwidth input (input forward) to a variable-bitwidth weight (from WBUF) to generate the product.
The \fusionunit then adds the product to an incoming partial sum to generate an outgoing partial sum (\texttt{Psum forward} in Figure~\ref{fig:bitfusion}(a)).

Figures~\ref{fig:bitfusion}(b),~\ref{fig:bitfusion}(c), and~\ref{fig:bitfusion}(d) show three different ways of logically fusing \bricks to form (b) 16 \fusedpes that support ternary (binary); (c) four \fusedpes that support mixed-bitwidths (2-bits for weights and 8-bits for inputs), (d) one \fusedpe that supports 8-bit operands, respectively.
For binary or ternary operations (Figures~\ref{fig:bitfusion}(b)), each \fusedpe contains a single \brick, offering the highest parallelism.
The \fusionunit then adds the results from all \fusedpes and the incoming partial sum to generate a single outgoing partial sum.
Figure~\ref{fig:bitfusion}(c) shows four \bricks fused together in a column to form a \fusedpe that can multiply 2-bit weights with 8-bit inputs.
The bitwidths of operands supported by a \fusedpe depend on the spatial arrangement of \bricks fused together.
Alternatively, by varying the spatial arrangement of the four fused \bricks, the \fusedpe can support 8-bit/2-bit, 4-bit/4-bit, and 2-bit/8-bit configurations for inputs/weights.
Finally, up to 16 \bricks can fuse together to construct a single \fusedpe that can operate on 8-bit operands for the multiply-add operations (Figure \ref{fig:bitfusion}(d)).
The \bricks fuse together in powers of 2. That is, a single \fusionunit with 16 \bricks can offer 1, 2, 4, 8, and 16 \fusedpes with varying operand bitwidths.
Dynamic composability of the \fusionunits at the bit level enables the architecture to expose the maximum possible level of parallelism with the finest granularity that matches the bitwidth of the DNN operands.

\begin{figure}
	\centering
	\includegraphics[width=0.8\linewidth]{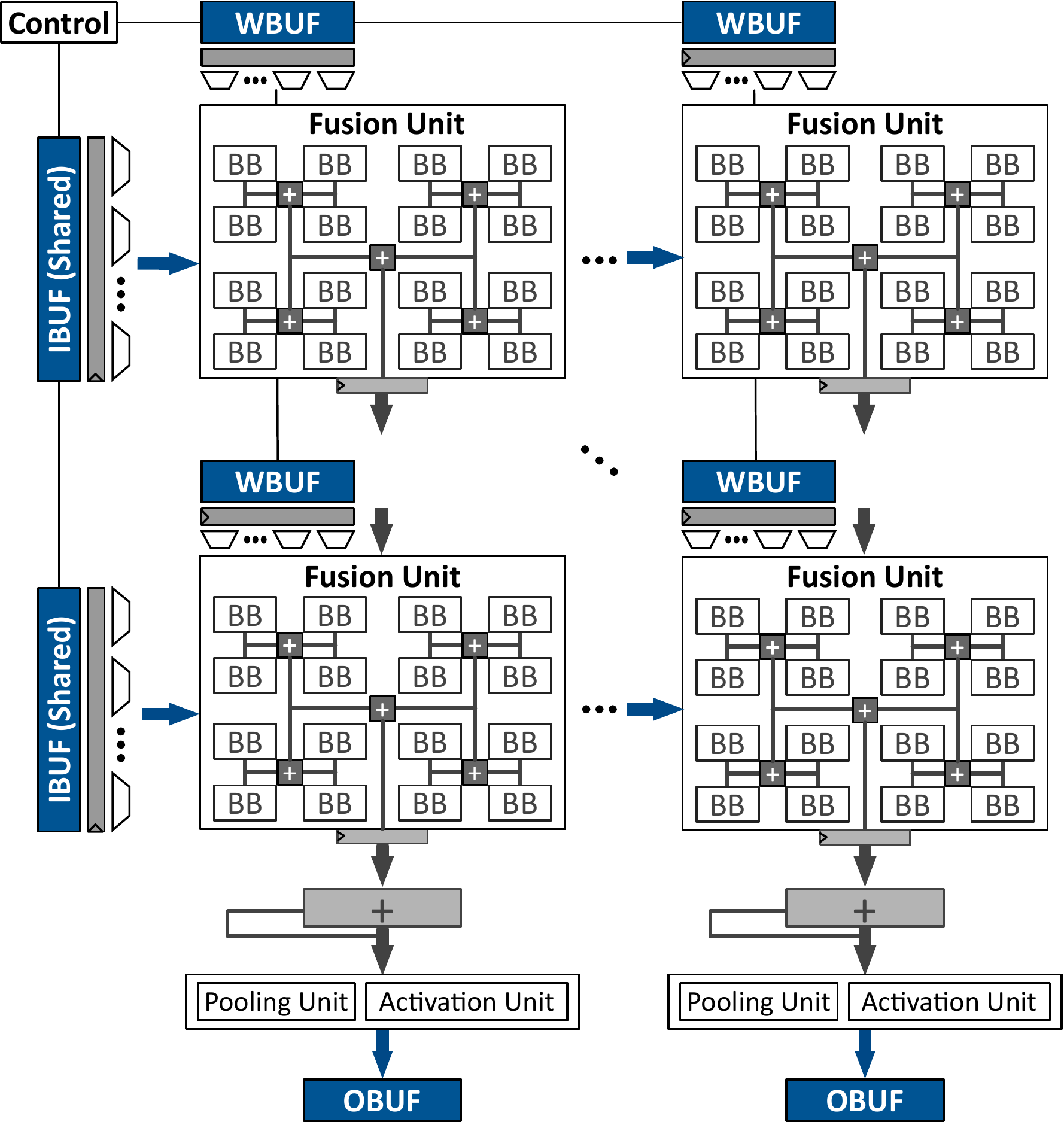}
	\caption{\bitfusion systolic architecture comprising a collection of \bricks (BBs) that can fuse to form \fusedpes.}
	\label{fig:arch}
	\vspace{-3.5ex}
\end{figure}

\begin{figure}
	\centering
	\includegraphics[width=0.7\linewidth]{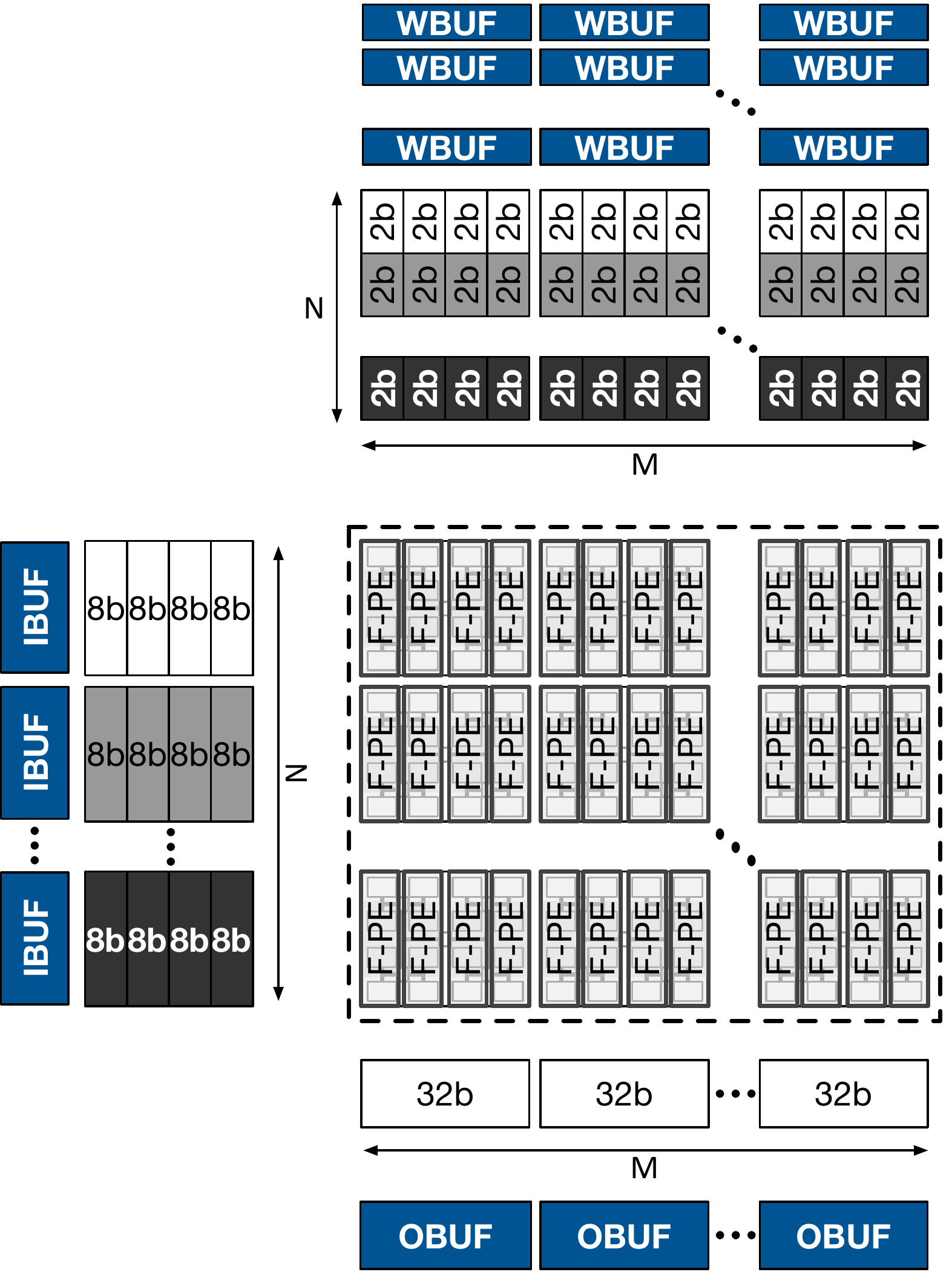}
	\vspace{-0.5ex}
	\caption{Bit-Flexible matrix-vector multiplication.}
	\label{fig:systolic-execution}
		\vspace{-1ex}
\end{figure}

\vspace{-1ex}
\subsection{Accelerator Organization}
Two insights guide the architecture design of \bitfusion.
First, DNNs offer high degrees of parallelism and benefit significantly from increasing the number of \fusionunits available within the accelerator's area budget.
Therefore, it is essential to minimize the overhead of control in the  accelerator by not only maximizing the number of \fusionunits but also minimizing the overhead of dynamically constructing \fusedpes, thereby integrating the maximum number of \bitbricks in the area budget.
Second, on-chip SRAM and register-file accesses dominate the energy consumption when accelerating DNNs~\cite{tetris:asplos:2017, eyeriss:isca:2016, eyeriss:jssc:2017}.
Therefore, it is essential to reduce the number of bits exchanged with on-chip and off-chip memory while maximizing data reuse.

\niparagraph{\bitfusion Systolic array.}
With these insights, we employ a 2-dimensional systolic array of \fusionunits as the architecture for \bitfusion, as shown in Figure~\ref{fig:arch}.
The systolic organization reduces the overhead of control by sharing the control logic across the entire systolic array.
More importantly, systolic execution alleviates the need for provisioning control for each \fusedpe as a dataflow architecture would have required.
As such, the systolic architectures fit the most number of \bitbricks in a given area budget.
Thus, the entire systolic array composed of \fusedpes acts as a single compute unit that can execute, for example, a single matrix-vector multiplication operation with various bitwidths, which also sets the level of parallelism.
In addition, the systolic organization of \fusionunits enforces sharing of input data across columns of the array and accumulates partial results across rows of the array to minimize access to on-chip memory.
As depicted in Figure~\ref{fig:arch}, the input buffers (\code{IBUFs}) only located at the borders and feed the rows simultaneously.
Similarly, the output buffers (\code{OBUFs}) reside on the bottom and collect the flowing results, which is accumulated by each column's accumulator.
As shown in Figure~\ref{fig:arch}, each column harbors a pooling and an activation unit before its output buffer.
Finally, the systolic organization also eliminates the need for local buffers for input, output, or partial results within \fusionunits.
As such, each \fusionunit is accompanied by only a weight buffer (\code{WBUF}).
Using \fusedpes as the building blocks, the performance of the systolic array maximally matches the bitwidths, with the highest performance at binary and ternary settings.

\begin{figure}
	\centering
	\includegraphics[width=0.8\linewidth]{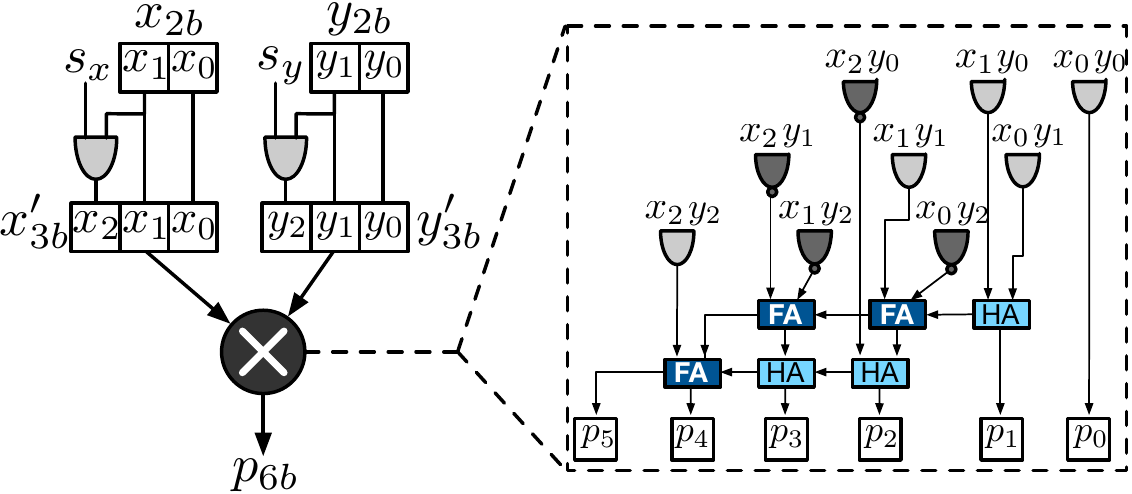}
	\vspace{-0.5ex}
	\caption{A single \brick. (HA: Half Adder, FA: Full Adder.)}
	\label{fig:bitbrick}
	\vspace{-3.5ex}
\end{figure}

\begin{figure*}[h]
\begin{minipage}{0.65\linewidth}
	\centering
	\hspace{-1ex}
	\includegraphics[height=2.2in]{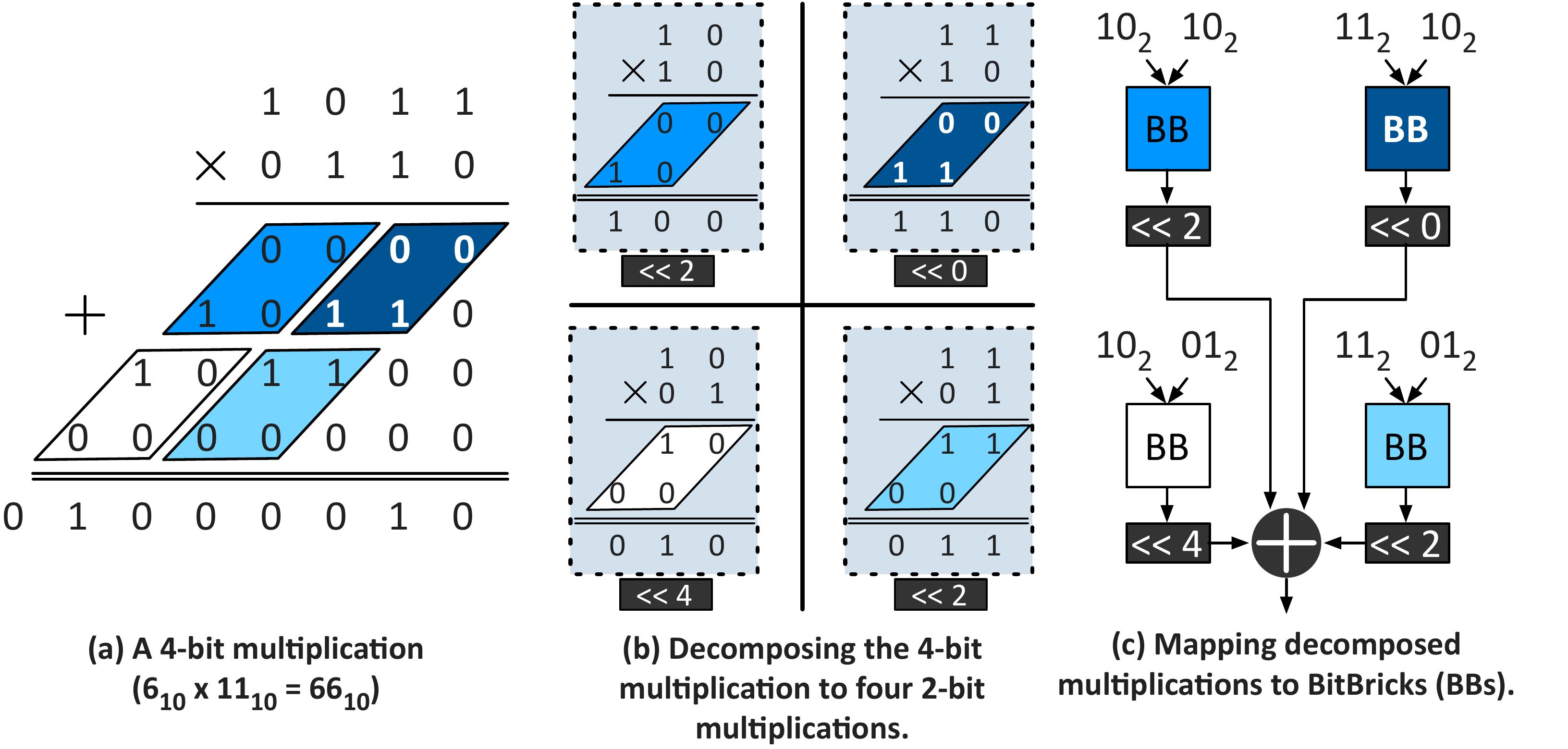}
	\caption{Using \bitbricks to execute $4$-bit multiplications.}
	\label{fig:bitbrick-4bx4b}
\end{minipage}
\begin{minipage}{0.3\linewidth}
	\centering
	\hspace{-1ex}
	\includegraphics[height=1.4in]{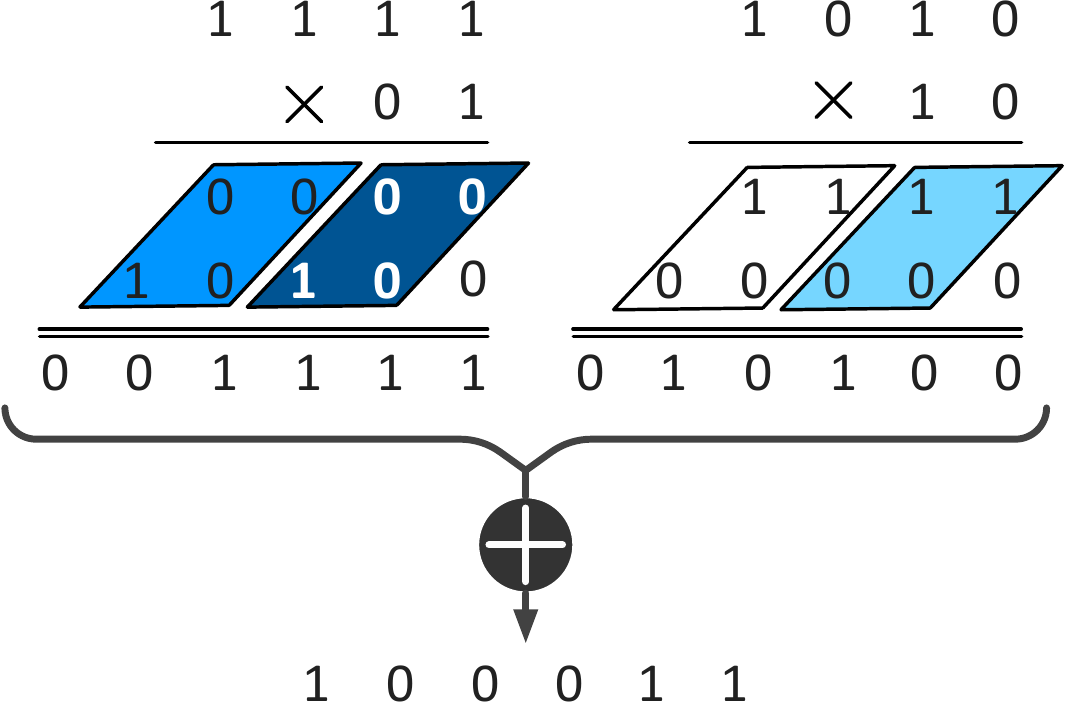}
	\caption{Two $4$-bit $\times$ $2$-bit multiplications decomposed to four $2$-bit multiplications followed by the accumulation (summation) logic.}
	\label{fig:bitbrick-4bx2b-algorithm}
\end{minipage}
\vspace{-3.5ex}
\end{figure*}

\niparagraph{Memory organization.}
Depending on the number of \fusedpes and their organization, the buffers must supply different number of operands with various bitwidths.
As such, we augment the input and the weight buffers with a register that holds a row of data that is gradually fed to the \fusedpes according to their bitwidth.
As illustrated in Figure~\ref{fig:arch}, a series of multiplexers after the register make this data infusion possible.
The benefit of this design is avoiding multiple accesses to the data array of the buffer which conserves energy.
With this design, at each cycle, the systolic array consumes a vector of inputs and matrix of weights to produce a vector of outputs with the fewest accesses to the buffers and the minimal bitwidth possible.

\vspace{-1ex}
\subsection{\bitfusionsubsection Execution Model}

Figure~\ref{fig:systolic-execution} illustrates the \bitfusion systolic execution in the mixed-bitwidth mode using when an input vector is multiplied to a weight matrix.
The input vector has $4\times N$ 8-bit elements that are being multiplied to a matrix with $4\times N \times M$ 2-bit elements.
As such, the 16-\bitbricks in a \fusionunit logically compose to form four 8$\times$2 \fusedpes.
Both input and weight buffers provide 32 bits per access.
The read values are split into 8-bit input values and 2-bit weight values in the output register of each buffer using its accompanying multiplexers as mentioned before.
The input values are shared across the \fusionunits of each row and weight values are specific to each \fusedpe.
As such, all of the $4\times N \times M$ \fusedpes work in parallel while only a single 32-bit value is read from the input and weight buffers.
Exploiting the lower bitwidth of weights, \bitfusion increases the level of parallelism by 4$\times$ while reducing the number of accesses to the weight buffer data arrays by the same factor of four.
As discussed above, each \fusionunit adds the results of its \fusedpes with its incoming partials results and forwards the partial output to the \fusionunit underneath it.
As shown in Figure~\ref{fig:systolic-execution}, we support 32-bit bitwidth for the partial and final results to avoid any inaccuracies.
\vspace{-1ex}
\section{Bit Fusion MicroArchitecture}
\label{sec:microarchitecture}

Given the overall organization of \bitfusion and its bit-flexible systolic execution model, this section delves into the details of \bitbricks and \fusionunits.
The key insight that enables bit-level dynamic composability in \bitfusion is the mathematical property that a multiply operation between operands with power-of-2 bitwidths ($4$-bit, $8$-bit, $16$-bit, and so on) can be decomposed to $2$-bit multiplications.
The products from the decomposed multiplications can then be put together by shift-add operations to generate the results of the original multiplication.
The bitwidths of the operands dictates the number of decomposed multiplications required and the shift amounts that are applied to the decomposed products before addition.
Using this insight, we design \bitbrick, the basic compute unit of the \bitfusion architecture, to support multiply operations for the smallest bitwidth of $2$-bits.
The $2$-bit operands for a \bitbrick can be both signed or unsigned.
Below, we describe the design of a single \bitbrick.

\vspace{-1ex}
\subsection{BitBrick Microarchitecture}

Figure~\ref{fig:bitbrick} shows the microarchitecture of a single \brick. 
As shown, a \bitbrick takes as input two 2-bit operands-- $x_{2b}$ and $y_{2b}$ and two corresponding sign-bits--$s_x$ and $s_y$.
The sign-bits $s_x$ and $s_y$ define if the 2-bit operands are signed (between -2 to 1) or unsigned (between 0 to 3).
According to the sign-bit, the \bricks first extend the 2-bit operands $x_{2b}$ or $y_{2b}$ to respectively create 3-bit sign extended operands $x'_{3b}$ or $y'_{3b}$.
Finally, the \bricks employ a 3-bit signed multiplier (shown with an encircled $\times$ in Figure~\ref{fig:bitbrick}) to generate a 6-bit product $p_{6b}$.
Thus, a \bitbrick supports both signed and unsigned numbers as its inputs.
The following subsection discusses how \bitfusion maps multiply-add operations with varying bitwidths to \bitbricks.

\vspace{-1ex}
\subsection{Mapping Variable Bitwidth Operations to \bitbricks}
To explain how \bitbricks compose to multiply operands with variable bitwidths, the discussion below uses a $4$-bit multiplication as an example.
As mentioned, a multiply operation with power-of-2 bitwidths can be decomposed to $2$-bit multiplies that can execute using \bitbricks.
Figure~\ref{fig:bitbrick-4bx4b}(a) illustrates this mathematical property for a multiplication between $4$-bit operands $1011_{2}$ $(11_{10})$ and $0110_{2}$ $(6_{10})$ to produce $01000010_{2}$ $(66_{10})$.
The $4$-bit multiplication in Figure~\ref{fig:bitbrick-4bx4b}(a) decomposes to four $2$-bit multiplications, shown in Figure~\ref{fig:bitbrick-4bx4b}(b).
The decomposed multiplications execute using \bitbricks to generate decomposed products, as shown in Figure~\ref{fig:bitbrick-4bx4b}(c).
The decomposed products require shifting before being put together.
For a $4$-bit multiplication using \bitbricks, the results from the decomposed $2$-bit multiplications are left-shifted by $0$, $2$, $2$, and $4$, as shown in Figure~\ref{fig:bitbrick-4bx4b}(c).

\niparagraph{Dynamic bitwidth flexibility.}
The bitwidths for the operands dictate how the results from the decomposed multiplications are left-shifted (multiplied with power of 2) before being added together.
By adding flexibility in the shifting logic, the \bitbricks can support $2$-bit and even mixed-bitwidth ($4$-bit $\times$ $2$-bit) multiplications.
Figure~\ref{fig:bitbrick-4bx2b-algorithm} shows the summation of two $4$-bit $\times$ $2$-bit multiplications ($15_{10}\times1_{10}$ $+$ $10_{10}\times2_{10}$ $=$ $35_{10}$).
The operation in Figure~\ref{fig:bitbrick-4bx2b-algorithm} breaks down to four $2$-bit decomposed multiplications that map to four \bitbricks.
Both the single $4$-bit $\times$ $4$-bit operation in Figure~\ref{fig:bitbrick-4bx4b}(a) and the two $4$-bit $\times$ $2$-bit operations in Figure~\ref{fig:bitbrick-4bx2b-algorithm} require the same number of \bitbricks.
Therefore, the performance at $4$-bit $\times$ $2$-bit is twice that of $4$-bit $\times$ $4$-bit.
The only difference between the operations in Figure~\ref{fig:bitbrick-4bx4b}(a) and Figure~\ref{fig:bitbrick-4bx2b-algorithm} is the shift amount required by the decomposed products.
Similarly, when operating at $2$-bit $\times$ $2$-bit, each \bitbrick can perform a single multiplication by setting the all the shift amounts to zero.

\begin{figure*}
\begin{minipage}{0.3\linewidth}
	\centering
	\includegraphics[height=2in]{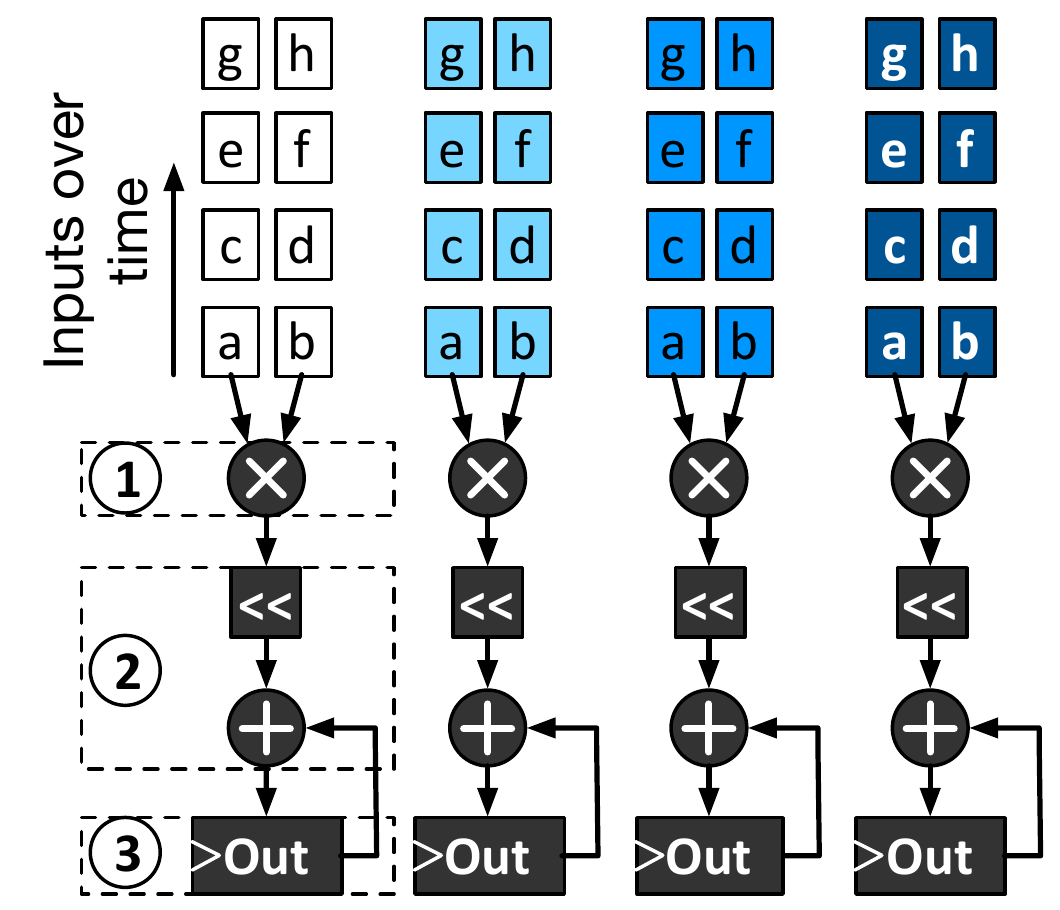}
	\caption{Temporal design. Operands $a-h$ are 2-bit.}
	\label{fig:temporal-mult}
\end{minipage}
\hspace{0.3in}
\begin{minipage}{0.3\linewidth}
	\centering
	\includegraphics[height=2in]{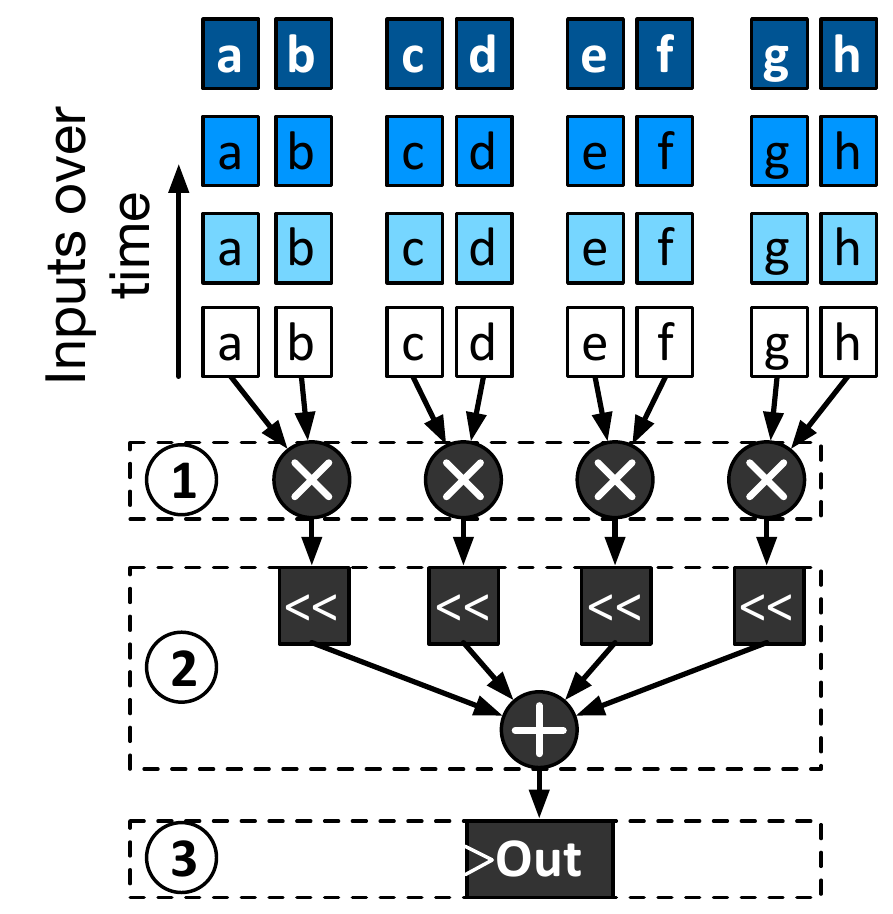}
	\vspace{-1ex}
	\caption{Spatial fusion. Operands $a-h$ are 2-bit.}
	\label{fig:spatial-mult}
\end{minipage}
\hspace{0.3in}
\begin{minipage}{0.3\linewidth}
	\centering
	\includegraphics[width=1.0\linewidth]{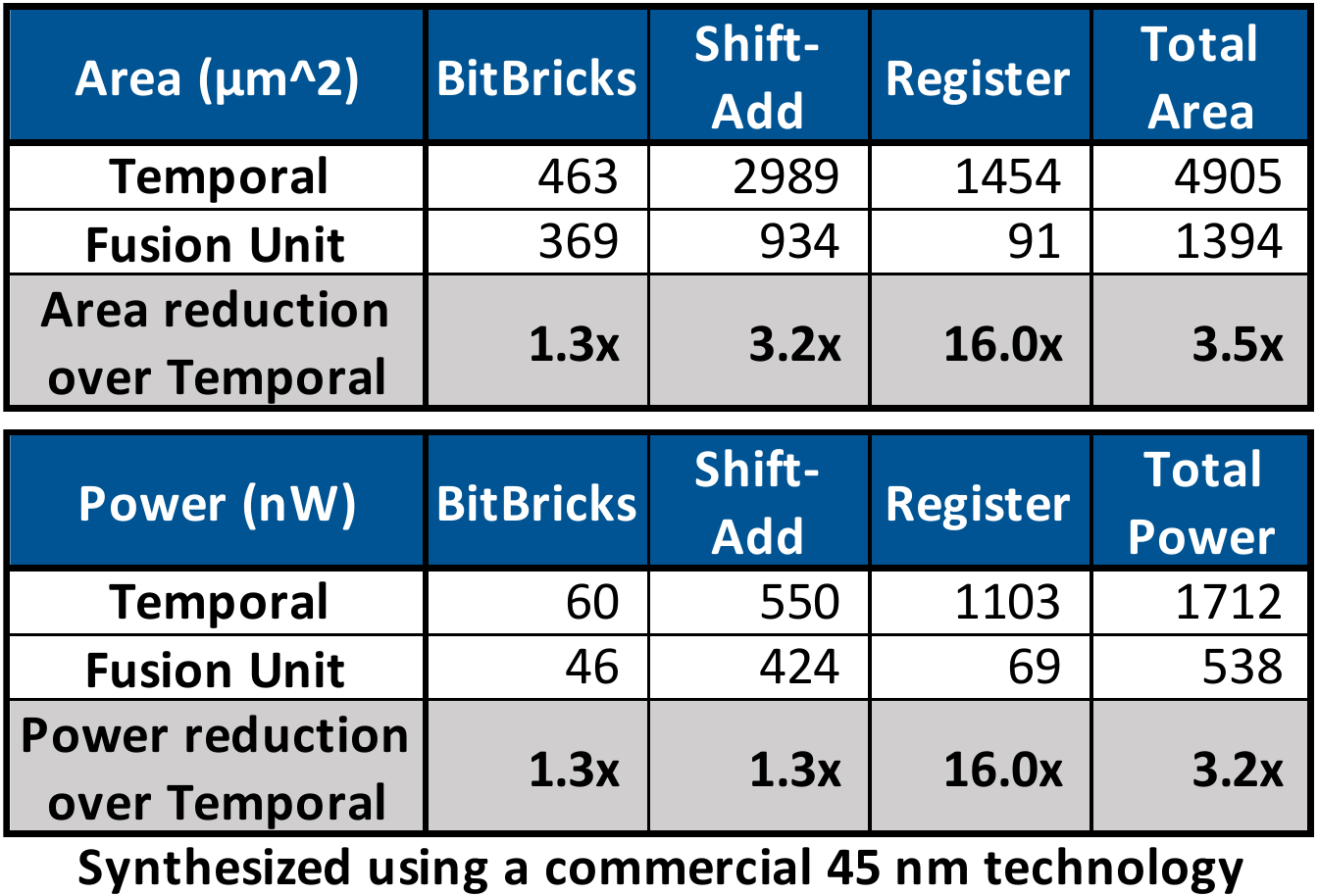}
	\caption{Area and Power comparison of the \fusionunit. Temporal design provided as reference.}
	\label{fig:fuse-mult-area-comparison}
\end{minipage}
\vspace{-3.5ex}
\end{figure*}

\niparagraph{Supporting arbitrary bitwidths.}
The discussion so far shows how multiply operations between $4$-bit and $2$-bit operands map to \bitbricks.
The same mathematical property can be recursively applied to support higher than $4$-bit for the operands.
\bitfusion supports up to $16$-bit operands by first recursively breaking down the $16$-bit multiplication to $8$-bit, $4$-bit and then $2$-bit multiplications which can execute using \bitbricks.
For a multiplication between $2n$-bit operands $A_{2n}$ and $B_{2n}$, the recursion can be expressed mathematically as follows.

\vspace{-2ex}
\begin{align}
	A_{2n} &= 2^{n} \times (A_{2n})_{hi} + 2^{0} \times (A_{2n})_{lo} \nonumber\\
	B_{2n} &= 2^{n} \times (B_{2n})_{hi} + 2^{0} \times (B_{2n})_{lo}
	\label{eqn:2n-n-breakdown}
	\\
	A_{2n} \times B_{2n}
		&= 2^{2n} \times (A_{2n})_{hi} \times (B_{2n})_{hi}
		+ 2^{n} \times (A_{2n})_{hi} \times (B_{2n})_{lo}
		\nonumber \\
		& + 2^{n} \times (A_{2n})_{lo} \times (B_{2n})_{hi}
		+ 2^0 \times (A_{2n})_{lo} \times (B_{2n})_{lo}
		\label{eqn:mult-2nx2n}
\end{align}

$(A_{2n})_{hi}$ and $(A_{2n})_{lo}$ refer to the $n$ most significant and $n$ least significant bits of $A$, respectively.
By applying the above equation recursively, \bitfusion supports up to $16$-bit operands.
When one of the operand's bitwidths is larger, we use the formulation below.

\vspace{-2ex}
\begin{align}
	A_{2n} \times & B_{n}
		= 2^{n} \times (A_{2n})_{hi} \times B_n
		+ 2^{0} \times (A_{2n})_{lo} \times B_n
		\label{eqn:mult-2nxn}
\end{align}

Each level of recursion, from $16$-bits to $8$-bits, $8$-bits to $4$-bits, and $4$-bits to $2$-bits, requires additional shift-add logic.
The overhead from the shift-add logic represents the hardware cost of bit-level flexibility.
The next subsection details the design of a \fusionunit that uses \bitbricks to execute multiply-adds with variable bitwidths, up to 16-bit.

\vspace{-1ex}
\subsection{\fusionunitsubsection Micro-Architecture}
To enable bit-level composability, \bitfusion introduces \emph{spatial fusion}, a paradigm that spatially combines the decomposed products generated by multiple \bitbricks over a single cycle.
Prior works~\cite{stripes:micro:2016, loom:arxiv:2017}, on the other hand, devise a temporal design that use single-bit multiply-add units independently over the span of multiple cycles.
The following elaborates on these two approaches. 
To offer a fair comparison, we assume that even the temporal design uses 2-bit multipliers, a configuration that provides a better area, delay, and power as opposed to a fully bit-serial design.

\niparagraph{Temporal design.}
Figure~\ref{fig:temporal-mult} shows a temporal design that can support variable bitwidths.
The variable-bitwidth multiply operation for the temporal design consists of three steps: (1) $2$-bit multiplication to generate a partial product, (2) shift operation to multiply with the appropriate power of $2$, and (3) accumulation in a register.
The temporal design requires $4$ cycles to execute a $4$-bit $\times$ $4$-bit multiplication.
The shift operation is simply a $4$-input multiplexer (mux).
Compared to a fixed $4$-bit multiplier, the temporal design uses much smaller multiply units for $2$-bit operands, which require significantly less area.
However, the number of gates required for the shifter and the accumulator depend on the highest supported bitwidth (16-bit for \bitfusion).
For instance, to support up to 16-bits using a temporal design, the shifter and the accumulator use up around $90\%$ of the area, which limits the benefits provided by this approach.
Nevertheless, the temporal design reduces area consumption over a fixed-bitwidth multiplier for the highest required bitwidth.

\niparagraph{Spatial fusion.}
In contrast, our spatial multiplier spatially combines (or fuses) the results from four \bitbricks over a single cycle to execute either one $4$-bit $\times$ $4$-bit multiplication, two $4$-bit $\times$ $2$-bit multiplications, or four $2$-bit $\times$ $2$-bit multiplications.
Figure~\ref{fig:spatial-mult} illustrates the design of a spatial multiplier that supports up to $4$ bits for either of the two operands using \bricks.
Similar to the temporal design, the spatial multiplier requires three steps: (1) multiplication using \bitbricks, (2) shift-add using the shift-add tree, and (3) accumulation of results in a register.
The spatial multiplier improves upon the temporal design by using a shift-add tree and a single shared accumulator to reduce the number of gates required.
Each level of the shift-add tree consists of three shift-units and a four-input adder that represent the multiplication with power of 2 in Equations (\ref{eqn:mult-2nx2n}) and (\ref{eqn:mult-2nxn}).
Compared to a $4$-bit fixed bitwidth multiplier the spatial multiplier requires more area but delivers $4\times$ higher performance for $2$-bit operations.
Overall, spatial fusion provides higher $\frac{performance}{area}$ compared to temporal design by packing more \bricks in the same area.

\niparagraph{\fusionunit using spatio-temporal fusion.}
As discussed, a \fusionunit can execute variable-bitwidth multiply-add operations and supports $2$-bit to $16$-bit operands.
Using Equations (\ref{eqn:mult-2nx2n}) and (\ref{eqn:mult-2nxn}) recursively, we can realize a \fusionunit using either the temporal design, spatial fusion, or a combination of both.
For a fixed area budget, using spatial fusion with 64 \bitbrick would pack the highest number of \bitbricks.
At the same time, feeding the 64 \bitbricks for spatial fusion would require $128$-bit wide accesses to the SRAM buffers (IBUF and WBUF in Figure~\ref{fig:arch}) per \fusionunit.
Increasing the width of SRAMs increases the area required by the IBUF and WBUF.
Therefore, we make a tradeoff wherein we use spatial fusion to combine $16$ \bitbricks spatially to realize support up to $8$-bit operands, and then combine it with temporal design to support up to $16$-bit operands over four cycles.
This hybrid approach balances both bit-level flexibility and the corresponding area overhead due to increased SRAM sizes.
Figure~\ref{fig:fuse-mult-area-comparison} compares the area and the power requirements for a \fusionunit with 16 \bitbricks that uses the hybrid approach with a temporal design using 16 \bitbricks.
As shown, for 16 \bitbricks, the hybrid \fusionunit has \xx{3.5$\times$} less area and \xx{3.2$\times$} less power compared to temporal design with the same number of 2-bit multipliers.

\niparagraph{Comparison to bit-serial temporal execution.}
%
Prior works in \stripes~\cite{stripes:micro:2016}, UNPU~\cite{unpu:isscc:2018}, and \loom~\cite{loom:arxiv:2017} devise bit-serial computation as a means to support flexible bitwidths for DNN operations.
%
Of the three, \loom is a fully-temporal architecture, similar to the temporal design discussed above (Figure~\ref{fig:temporal-mult}).
%
\stripes and UNPU are hybrid designs that fix the bitwidth of one operand and support variable bitwidths for the other.
%
We provide a head-to-head comparison to \stripes in Section~\ref{subsubsec:stipes-comparison} and provide a qualitative comparison to \loom below.
As the results from Figure~\ref{fig:fuse-mult-area-comparison} indicate, for the same throughput, a fully-temporal design, such as the one used in \loom, would consume significantly larger area and power compared to our spatially composable \fusionunit.
Furthermore, a fully-temporal design iterates in the form of a nested loop over the bits the two operands; hence, requiring more number of accesses to the SRAM.

The next section discusses the \bitfusion-ISA, that exposes the bit-level flexibility of \bitfusion to software.
\vspace{-1ex}
\section{Instruction Set Architecture}
\label{sec:isa}

\begin{table}
	\centering
	\caption{\bitfusion Instruction Set.}
	\vspace{-1ex}
	\includegraphics[width=0.92\linewidth]{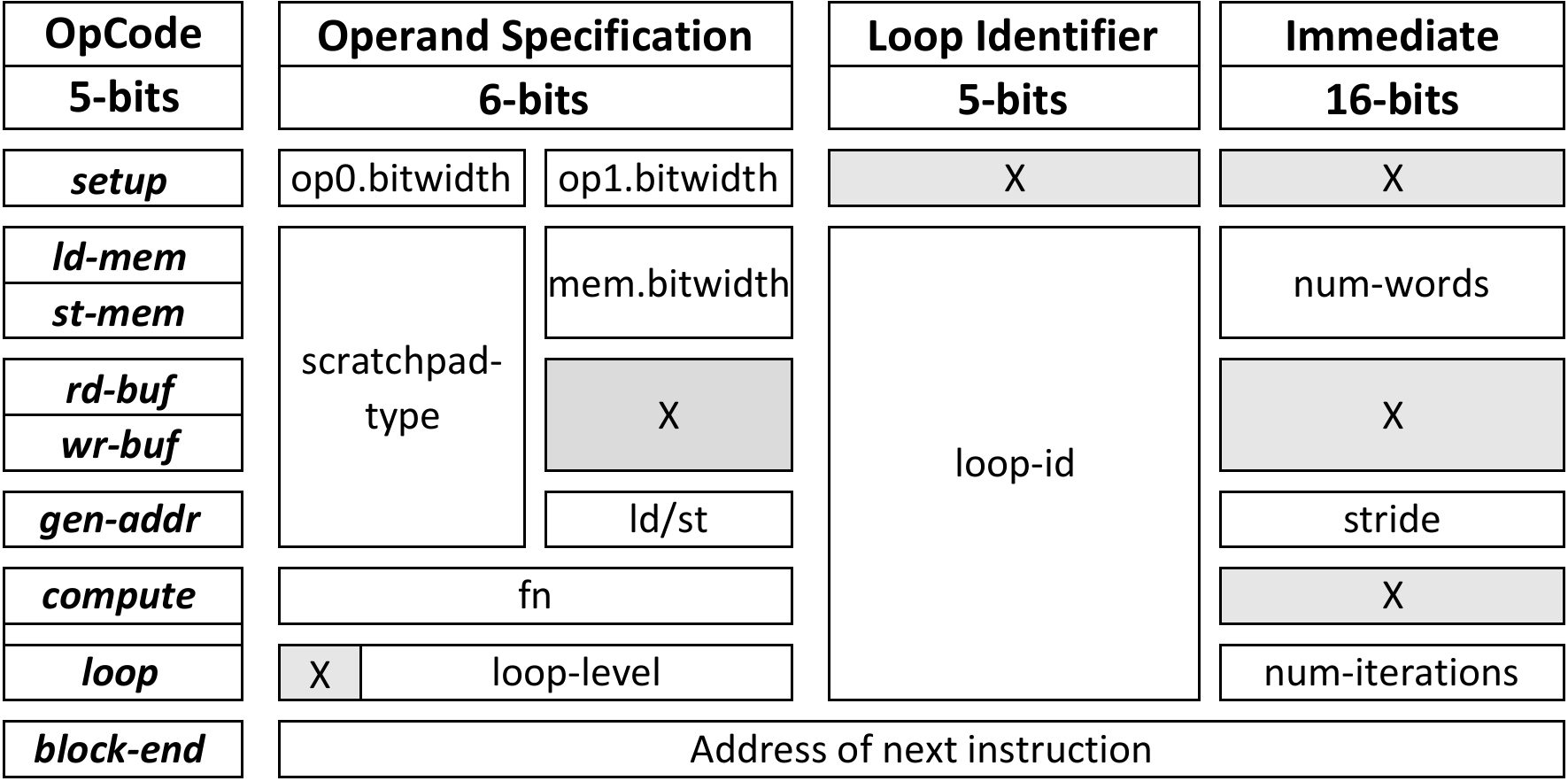}
	\label{tab:isa}
	\vspace{-3.5ex}
\end{table}

To leverage the unique bit-level flexibility of \bitfusion, we need to design a new hardware-software interface that exposes those capabilities in an abstract manner. 
Furthermore, the abstraction must be flexible to enable a wide range of DNN models so as to exploit bit-level fusion.
The following lists the requirements for an ISA that provides this abstraction and enables efficient use of \bitfusion for various categories of DNNs.
\begin{enumpacked}
	\item \textbf{Amortize the cost of bit-level fusion by grouping operations.} The operations in a DNN are organized into groups, called layers, wherein the same mathematical operation repeats a large number of times (often hundreds of thousands). To avoid the overhead of \emph{fine-grained} control over the operations at such a scale, the abstraction needs to amortize the cost of bit-level fusion across blocks of instruction that implement the layers.
	
	\item \textbf{Enable a flexible data-path for \bitfusion.} Both the number of words and the bitwidth of each word that feeds the \fusedpes varies depending on how the \bricks are composed as discussed in Section~\ref{sec:arch}. Thus, the semantics of instructions for data accesses must vary according to the fusion configuration to enable a flexible data-path.	
			
 	\item \textbf{Provide a concise expression for a wide range of DNN layers}. As research in DNNs is still volatile, it is necessary to devise an ISA that is general enough to express a wide range of DNN operations/layers. Yet, minimizes the von Neumann overhead of instruction handling and require a small footprint.
\end{enumpacked}

\vspace{-1ex}
\subsection{\fusionisasubsection for Bit-Flexible Acceleration}
Table~\ref{tab:isa} summarizes the \bitfusion-ISA that aims to satisfy these requirements.
The rest of this section discusses the instruction formats and provides the insight that drives them.

\niparagraph{Block-structured ISA for DNN layers.}
To leverage the commonalities in the operations of a layer, the \bitfusion ISA is \emph{block structured}.
As such, the fusion configuration of the \bitbricks is fixed across each block of instructions that implement a specific layer.
In this work, we did not explore within layer bitwidth variations. 
Nevertheless, the \bitfusion ISA and this incarnation of its microarchitecture can readily support it by using multiple instruction blocks for an individual layer.
The \code{setup} instruction marks the beginning of an instruction block and configures the \fusionunits and its data delivery logic to the specified bitwidth for the operands.
This instruction effectively defines the \emph{logical} fusion of the \bricks into \fusedpes for all the instructions in the block.
The \code{block-end} instruction signifies the end of a block and provides the address to the next instruction in the \code{next-inst} field.

\niparagraph{Concise expression of DNN layers.}
DNNs consist of a large number of simple operations like multiply-accumulate and max, repeated over a large number of neurons (over 2600 million multiply-adds  in AlexNet. See Table~\ref{tab:benchmarks}).
Thus, the von Neumann overhead of instruction fetch and decode can limit performance due to the large number of operations required by a DNN.
To minimize the number of instruction fetches/decodes required, we leverage the following insight.
Each layer in a DNN is a series of simple mathematical operations over hyper-dimensional arrays.
How the operations walk through the array elements and the type of mathematical operation (multiply-add/max) uniquely defines a layer.
As such, the ISA provides \code{loop} instructions that enable a concise way of defining the walks and operations in a DNN layer.
Each \code{loop} instruction has a unique ID in the block.
As shown in Table~\ref{tab:isa}, the \code{num-iterations} field in the \code{loop} instruction defines iteration count.
The \code{compute} instruction specifies the type of operation, while the \code{gen-addr} instruction dictates how to walk through the elements of the input/output hyper-dimensional arrays.
The \code{stride} field in the \code{gen-addr} instruction specifies how to walk through the array elements in the \code{loop}, which is identified by the \code{loop-id} field.
The words after the \code{setup} instruction define the memory base address for the data that fills the three buffers of input, output, and weights.
The \code{gen-addr} instruction generates the addresses that walk through the memory data and fill the buffers.
\begin{equation}
	Address = base + \sum_{id} (loop\_iterator[id] \times stride[id])
	\label{eq:addr}
\end{equation}

In Equation~(\ref{eq:addr}), $id$ is the \code{loop-id} field of all the \code{gen-addr} instruction in the block and the $loop\_iterator$ is the current iteration of the corresponding loops and their $stride$s.
The fundamental assumption is that multiple \code{gen-addr} instructions repeated by corresponding \code{loop} instructions define the complex multi-dimensional walks that expresses various kinds of DNN layers from LSTM to CNN.
In the evaluated benchmarks, blocks with \xx{30}-\xx{86} instructions are enough to cover LSTM, CNN, pooling, and fully connected.
These blocks use a combination of \code{loop}, \code{compute}, and \code{gen-addr} instructions to define these DNN layers nested loops.
These statistics show that our ISA can concisely express various DNN layers while providing bit-level fusion capabilities.
Note that these instructions are fetched and decoded once at the beginning of an instruction block, amortizing the von Neumann overhead over the entire execution of the block.

\niparagraph{Managing memory accesses for \fusedpes.}
The \code{ld-mem}/\code{st-mem} instructions exchange data between the on-chip buffers (\code{IBUF}, \code{OBUF}, and \code{WBUF}) in Figure~\ref{fig:arch}--and the off-chip memory.
Similarly, the \code{rd-buf}/\code{wr-buf} instructions read/write data from the on-chip buffers specified by the \code{scratchpad-type} field as shown in Table~\ref{tab:isa}.
In these four instructions, the size of the operands, which are variable-bitwidth arrays, depends on the number of array elements and their bitwidths.
These parameters, which control the logic that feeds the \fusedpes, are dependent on the bit-level fusion configuration (number of \fusedpes in each \fusionunit) and the type of data (input/weights).
To capture this variation in the size of data, the semantics of \code{rd-buf}/\code{wr-buf} and \code{ld-mem}/\code{st-mem} instructions for accessing on-chip and off-chip memory vary according to the fusion configuration of their instruction block, set apriori.
In particular, the sizes of memory accesses by \code{ld-mem}/\code{st-mem} instructions depend on both its \code{num-words} field and the fusion configuration defined by the corresponding \code{setup} instruction.

\niparagraph{Decoupling on-chip and off-chip memory accesses.}
The data required by DNNs, and subsequently, the number of memory accesses are large.
Hence, the latency due to off-chip memory accesses can be a performance bottleneck.
To hide the latency of off-chip accesses, the ISA decouples the on-chip memory accesses with off-chip.
Furthermore, decoupling the two types of memory accesses allows the accelerator to reuse on-chip data using simple scratchpad buffers, instead of hardware-managed caches.
\vspace{-1ex}
\subsection{Code Optimizations}
\label{sec:compiler-opt}

\begin{figure}
	\centering
	\includegraphics[width=0.55\linewidth]{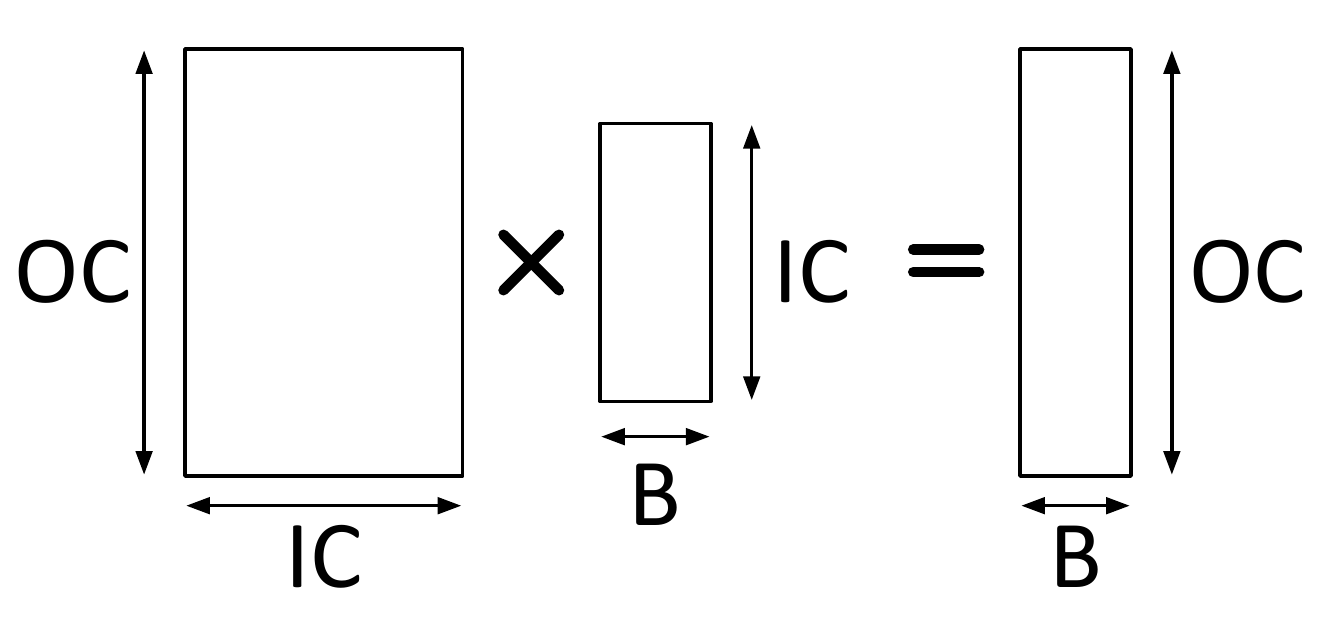}
	\vspace{-1ex}
	\caption{A single Fully-Connected Layer. The $\times$ represents matrix multiplication.}
	\label{fig:compiler-example-network}
	\vspace{-1ex}
\end{figure}

\begin{figure}
	\centering
	\includegraphics[width=0.9\linewidth]{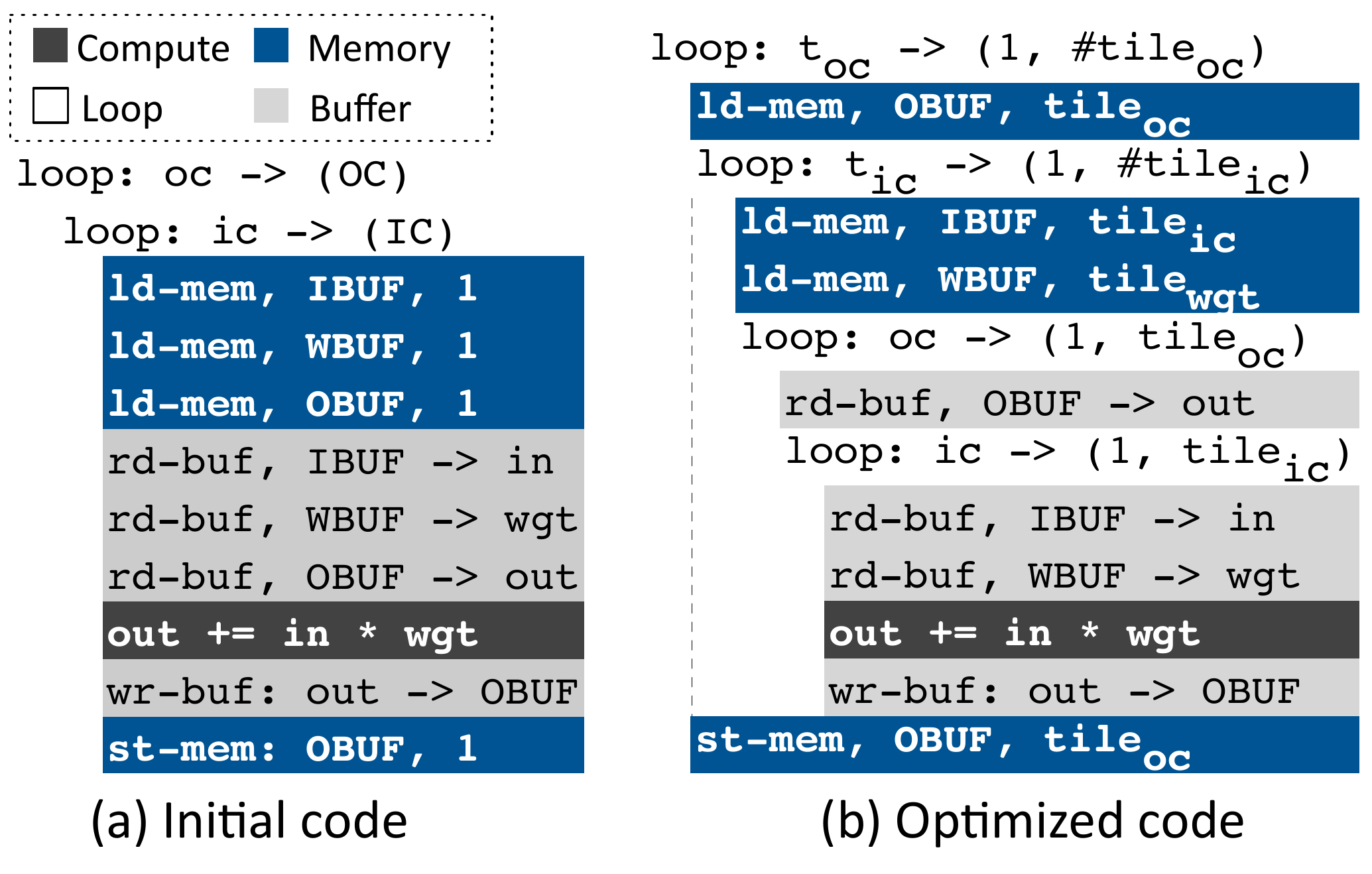}
	\vspace{-1ex}
	\caption{(a) Code for the Fully-Connected Layer. (b) Optimized code using loop tiling and ordering. \code{setup} and \code{gen-addr} instructions omitted for clarity.}
	\label{fig:compiler-instructions}
	\vspace{-3.5ex}
\end{figure}

As discussed in Section~\ref{sec:isa}, the \fusionisa uses simple instructions combined with explicit loop instructions to express neural networks.
The use of simpler instructions makes the ISA flexible to express a large range of DNNs.
Nonetheless, the flexibility in the ISA enables incorporating layer-specific optimizations to improve the performance and energy gains.
For brevity, we use an example fully-connected layer to discuss the code optimizations. Figure~\ref{fig:compiler-example-network} shows the matrix-matrix multiplication associated with this example.
We perform the following three optimizations as depicted in Figure~\ref{fig:compiler-instructions}.

\niparagraph{Loop ordering.}
Loop-ordering optimizes the order of the outer loops and memory instructions to further reduce off-chip accesses.
Recall that \bitfusion-ISA uses loop indices to generate memory addresses (Section~\ref{sec:isa}).
When the address for a memory instruction does not depend on the index of the previous loop instruction, their order can be exchanged.
The optimized code in Figure~\ref{fig:compiler-instructions}(b) uses \code{Output-Stationary} for executing the fully-connected layer, to reduce read/write accesses to the output buffer.
Changing the order allows \bitfusion to switch between \code{Input-Stationary}, \code{Output-Stationary}, and \code{Weight-Stationary} to minimize off-chip and on-chip accesses.

\niparagraph{Loop tiling.}
Loop-tiling partitions a loop instruction in the \bitfusion-ISA into smaller \emph{tiles} such that the data required by a loop operation fits inside the on-chip scratchpads.
The smaller tiles are accessed using a single LD/ST instruction and are reused in the inner-loop to reduce off-chip accesses.
Compared to the original code in Figure~\ref{fig:compiler-instructions}(a), the tiled version in Figure~\ref{fig:compiler-instructions}(b) reduces off-chip accesses for output buffer by a factor of \code{IC$\times$}, and on-chip accesses for output buffer by a factor of \code{$tile_{ic}$}. Note that \code{IC} is a  dimension in the matrix multiplication operation as depicted in Figure~\ref{fig:compiler-example-network}.
Convolution layers typically require six loop instructions, which increases to 12 after tiling optimizations.
The overhead of increasing the number of instructions on performance is negligible since the cost of fetch and decode is amortized throughout the execution of the layer.

\niparagraph{Layer fusion.}
As discussed, the \bitfusion architecture consists of a 2-D systolic array of multipliers, along with a 1-D array of pooling/activation units.
When two or more consecutive layers use mutually exclusive on-chip resources, the instructions for the two layers are combined such that the data produced by the first layer is directly fed into the subsequent layer, avoiding costly off-chip accesses.
For example, the fully-connected layer in Figure~\ref{fig:compiler-example-network} uses the 2-D systolic array.
If the next layer is activation, then we can fuse the layers and create one block of instruction for computing both the layers.
\vspace{-1ex}
\section{Evaluation}
\label{sec:eval}

\vspace{-0.5ex}
\subsection{Methodology}
\label{sec:methodology}

\niparagraph{Benchmarks.}
Table~\ref{tab:benchmarks} shows the list of 8 CNN and RNN benchmarks from diverse domains including image classification, object and optical character recognition, and language modeling.
The selected DNN benchmarks use a diverse size of input data, which allows us to evaluate the effect of input data size on the \bitfusion architecture.
\bench{AlexNet}~\cite{alexnet, wrpn}, \bench{SVHN}~\cite{svhn:nips:2011, qnn:arxiv:2016}, \bench{CIFAR10}~\cite{cifar10, qnn:arxiv:2016}, \bench{LeNet-5}~\cite{lenet-5, li2016ternary}, \bench{VGG-7}~\cite{vgg, li2016ternary}, \bench{ResNet-18}~\cite{resnet, wrpn} are popular and widely-used CNN models.
Among them, \bench{AlexNet} and \bench{ResNet-18} benchmarks are image classification applications that have different network topologies that use the ImageNet dataset.
The \bench{SVHN} and \bench{LeNet-5} benchmarks are optical character recognition applications that recognize the house numbers from the house view photos and handwritten/machine-printed characters, respectively.
\bench{CIFAR10} and \bench{VGG-7} are object recognition applications based on the CIFAR-10 and ImageNet dataset, respectively.
The \bench{RNN}~\cite{qnn:arxiv:2016} and \bench{LSTM}~\cite{lstm, qnn:arxiv:2016} are recurrent networks that perform language modeling on the Penn TreeBank dataset~\cite{penn-treebank}. 
In Table~\ref{tab:benchmarks}, the ``Multiply-Add Operations'' column shows the required number of Multiply-Add operations for each model and the ``Model Weights'' column shows the size of model parameter.
Note that the multiply-add operations and model weights have variable bitwidths as presented in Figure~\ref{fig:motivation}.

\niparagraph{Reduced bitwidth DNN models.}
\bitfusion aims to accelerate the inference of a wide range of DNN models with varying bitwidth requirements, with \emph{no loss in classification accuracy}.
The benchmarks, listed in Table~\ref{tab:benchmarks}, employ the model topologies proposed in prior work~\cite{dorefa:arxiv:2016, qnn:arxiv:2016, li2016ternary, wrpn} that train low bitwidth DNNs and achieve the same accuracy as the 32-bit floating-point models.
We did not engineer these quantized DNNs and merely took them from the existing  deep learning literature~\cite{dorefa:arxiv:2016, qnn:arxiv:2016, li2016ternary, wrpn}.
Benchmarks \bench{Cifar-10}, \bench{SVHN}, \bench{LSTM}, and \bench{RNN} use the quantized models presented in~\cite{qnn:arxiv:2016}.
Benchmarks \bench{LeNet-5} and \bench{VGG-7} use ternary (+1,0,-1) networks~\cite{li2016ternary}.
\bench{AlexNet} and \bench{ResNet-18} use the $4$-bit $2\times$ wide models presented in~\cite{wrpn} that double the number of channels for convolution and fully-connected layers.
We use the regular \bench{AlexNet} and \bench{ResNet-18} models for \eyeriss and the GPU baselines, and use their $2\times$ wide quantized models for \bitfusion and \stripes.

\begin{table}
	\centering
	\caption{Evaluated CNN/RNN benchmarks.}
	\vspace{-1ex}
	\includegraphics[width=0.95\linewidth]{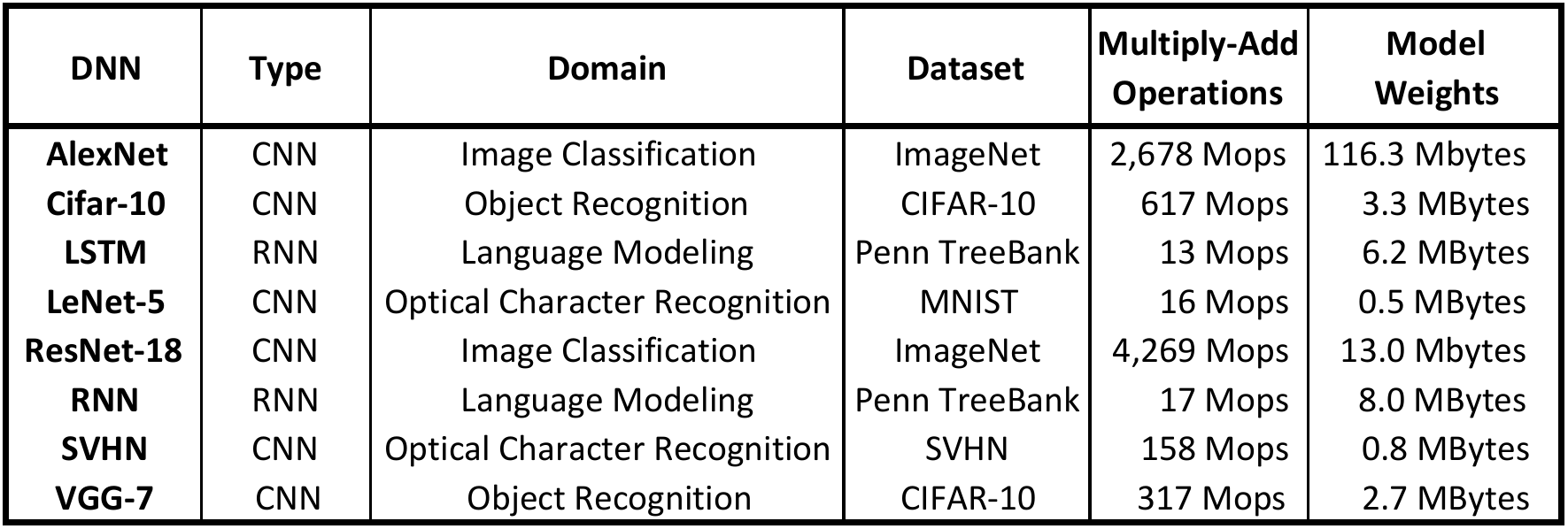}
	\vspace{-1ex}
	\label{tab:benchmarks}
\end{table}

\begin{table}
	\centering
	\caption{Evaluated ASIC and GPU platforms. *\stripes entries per-tile.}
	\vspace{-1ex}
	\includegraphics[width=0.95\linewidth]{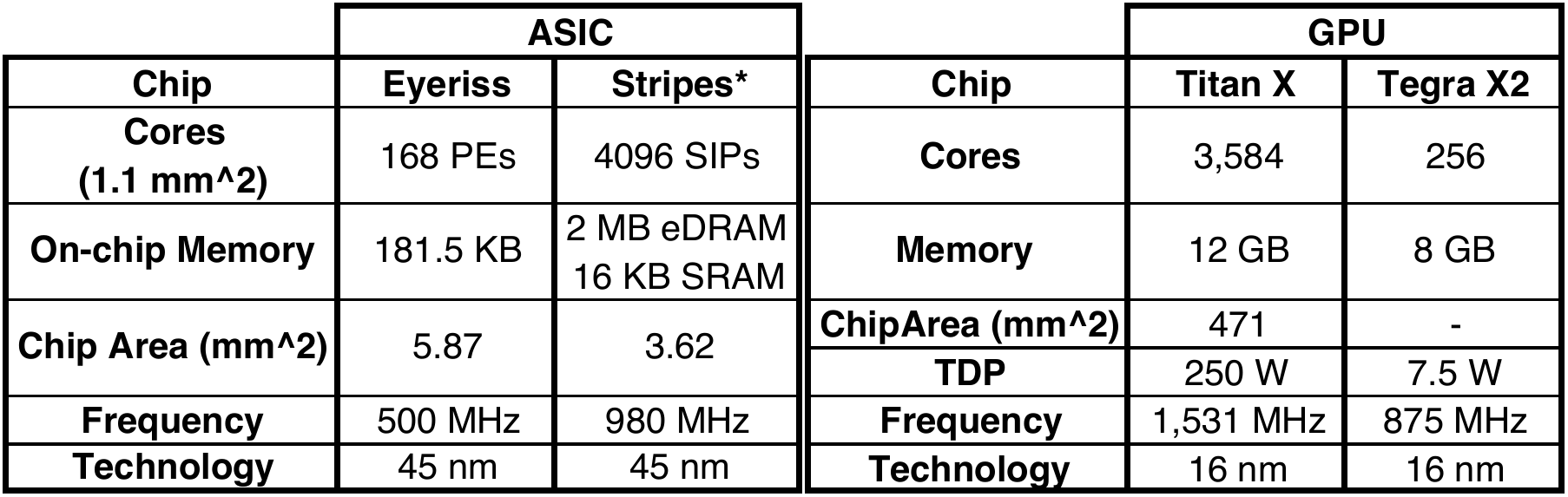}
	\vspace{-3ex}
	\label{tab:platforms}
\end{table}

\niparagraph{Accelerator development and synthesis.}
We use RTL-Verilog to implement the configuration of the \bitfusion architecture and verify the design through extensive RTL-simulations.
We synthesize \bitfusion at 45nm technology node using \bench{Synopsys Design Compiler (L-2016.03-SP5)} and a commercial standard-cell library.
\bench{Design Compiler} provides the chip area, achievable frequency, and dynamic/static power, which we use to estimate the performance and energy-efficiency of the \bitfusion accelerator.

\niparagraph{Simulation infrastructure for \bitfusion.}
We compile each DNN benchmark to the instructions of the \fusionisa (Section~\ref{sec:isa}).
We develop a cycle-accurate simulator that takes the \fusionisa instructions for the given DNN and simulates the execution to calculate the cycle counts as well as the number of accesses to on-chip buffers (IBUF, OBUF, and WBUF in Figure~\ref{fig:arch}) and off-chip memory.
We verify the cycle counts of the simulator against our Verilog implementation of the \bitfusion architecture.
Using the frequency defined in Table~\ref{tab:platforms} and the cycle counts, the simulator measures the execution time of the \bitfusion architecture.
To evaluate the energy efficiency, we model the energy consumption for on-chip buffers for the \bitfusion accelerator using the results from CACTI-P~\cite{cactip}.

\niparagraph{Comparison with Eyeriss.}
To measure the performance and energy dissipation of our comparison point, \eyeriss, we use their open-source simulation infrastructure~\cite{tetris:asplos:2017}.
The resulting area and energy metrics are shown in Table~\ref{tab:platforms}.
As mentioned, we use the same area budgets as \eyeriss, which is 1.1 mm$^2$ for compute units and 5.87 mm$^2$ for chip to synthesize \bitfusion, shown in Table~\ref{tab:platforms}.
We use a total 112~KB SRAM for on-chip buffers (IBUF, OBUF, and WBUF in Figure~\ref{fig:arch}).
Eyeriss operates on the 16-bit operands and \bitfusion supports flexible bitwidths from 2, 4, 8, to 16 bits.

\niparagraph{Comparison with Stripes.}
The authors of Stripes graciously shared their simulator~\cite{stripes:micro:2016}.
Their power estimation tools were in 65~nm node, which we scaled to 45~nm.
\stripes operates on 16-bit inputs and variable-bitwidth weights (1 through 16), using Serial Inner-Product units (SIPs).
\stripes is organized into 16 tiles each of which has \xx{4096} SIPs.
For a fair comparison, we replace the \xx{4096} SIPs in each tile of \stripes with our proposed \bitfusion systolic array with 512 \fusionunits, each with 16 \bricks to match the same budget of \xx{1.1mm$^2$} for compute, which is the area after scaling to 45~nm and use the same total on-chip memory.

\niparagraph{Comparison with GPUs.}
We use two GPUs (\titanxp and Tegra X2) based on Nvidia's Pascal architecture to compare with \bitfusion.
Table~\ref{tab:platforms} shows the details of the two GPUs.
We use Nvidia's custom \bench{TensorRT 4.0}~\cite{tensorrt} library compiled with the latest CUDA~\xx{9.0} and cuDNN~\xx{7.1} which support 8-bit quantized calculations, the smallest possible in the architecture.
Across GPU platforms, we use 1,000 warm-up batches, followed by 10,000 batches to measure performance and use the average.
For a head-to-head comparison, we conservatively scale \bitfusion to 16~nm technology node assuming a \xx{$0.86\times$} voltage scaling and \xx{$0.42\times$} capacitance scaling according to the methodology presented in~\cite{dark_silicon:isca}.
However, we assume the same frequency of \xx{500 MHz} as \eyeriss and do not increase the \bitfusion frequency.
The scaled \bitfusion architecture has 4096 \fusionunits with \xx{896 KB} SRAM and has a total chip area of \xx{5.93 mm$^2$} and consumes 895~milliwatts of power.
As a point of reference, \titanxp in the same 16~nm node, has a chip area of 471~mm$^2$ and has a TDP of 250~Watts, as summarized in Table~\ref{tab:platforms}.
\vspace{-1ex}
\subsection{Experimental Results}

\subsubsection{Comparison to \eyeriss}

\begin{figure}
	\centering
	\includegraphics[width=1.0\linewidth]{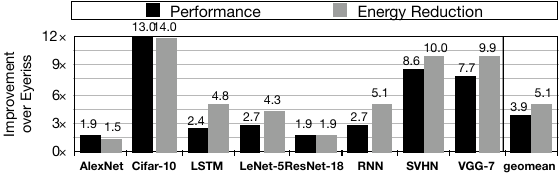}
	\vspace{-3ex}
	\caption{\bitfusion performance and energy improvements over \eyeriss.}
	\vspace{-3.5ex}
	\label{fig:improvement-over-eyeriss}
\end{figure}

\niparagraph{\\Performance and energy improvement.}
To evaluate the performance and efficiency benefits from the \bitfusion architecture, we compare with a state-of-the-art accelerator \eyeriss~\cite{eyeriss:isca:2016} that proposes an optimized dataflow architecture for DNNs.
We match the same area budget of \xx{1.1mm$^2$} for computational logic across both architectures: systolic array in \bitfusion and PEs in \eyeriss, and match the total SRAM capacity.
We scale the area and energy consumption of the PEs, register-files, on-chip network, and DRAM in \eyeriss to 45nm technology according to the methodology proposed in~\cite{tetris:asplos:2017}.
For a fair comparison between the two architectures, we use the same frequency of \xx{500MHz} reported in the paper~\cite{tetris:asplos:2017} for both \eyeriss and \bitfusion.
Figure~\ref{fig:improvement-over-eyeriss} presents the performance and energy benefits of \bitfusion in comparison with \eyeriss.
On average, \bitfusion delivers \eyerissPerfAvg speedup since the \bitfusion architecture can perform more DNN operations with lower bitwidth in a given area compared to \eyeriss.
Depending on the types of DNN operations and the required bitwidths, the benchmarks see different performance gains.
The CNN benchmarks (\bench{AlexNet}, \bench{SVHN}, \bench{Cifar-10}, \bench{LeNet-5}, \bench{VGG-7}, and \bench{ResNet-18}) see higher performance gains than the recurrent networks (\bench{RNN} and \bench{LSTM}) since the convolution operations are more amenable for data reuse in systolic architecture of \bitfusion.
\bench{Cifar-10} sees the highest benefits of \xx{13$\times$} speedup since most of its operations can be computed with the smallest bitwidth (1-bit input and 1-bit weight) and its operations provide a large degree of parallelism that can exploit the increased number of \fusedpes.
In contrast, \bench{ResNet-18} and \bench{AlexNet} achieve the lowest speedup of \xx{1.9$\times$}, because these two benchmarks use twice the number of channels ($2\times$ wide) for convolution and fully-connected layers~\cite{wrpn} for quantized execution on \bitfusion.
We use the original \bench{AlexNet} and \bench{ResNet-18} models on \eyeriss, which effectively requires $4\times$ less multiply-add operations.
Overall, using variable bitwidth improves performance and energy efficiency, since it increases compute capacity and reduces active hardware components.
Figure~\ref{fig:improvement-over-eyeriss} also shows the energy reduction.
The average improvement is \eyerissEnergyAvg, with the largest of \xx{14$\times$} from \bench{Cifar-10} and the smallest of \xx{1.5$\times$} from \bench{AlexNet}.
The significant energy reduction attributes to both \fusionunit organizations and memory access reductions, which we discuss below in more detail.

\niparagraph{Energy breakdown.}
\begin{figure}
	\centering
	\includegraphics[width=0.95\linewidth]{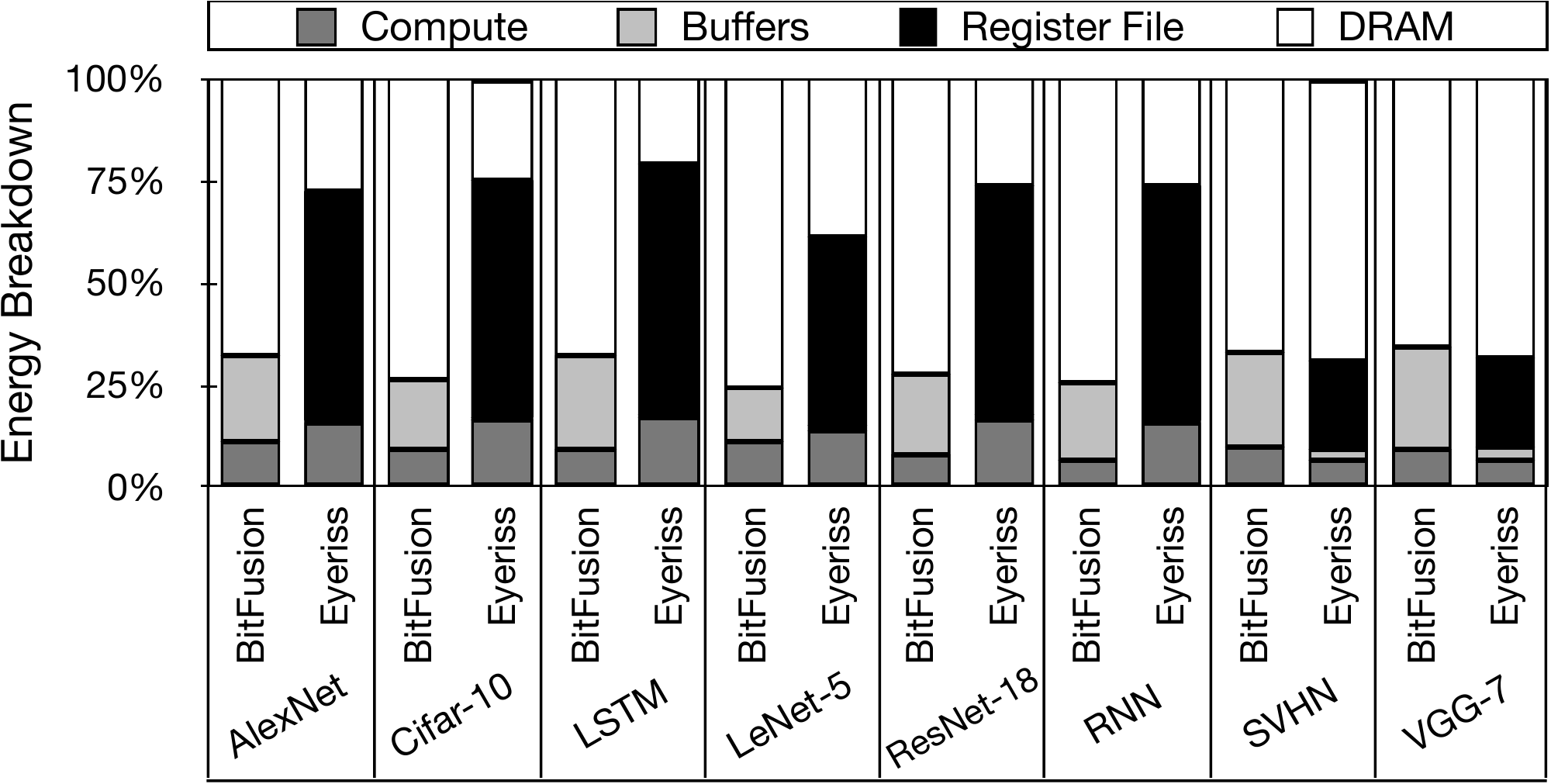}
	\caption{Energy breakdown of \bitfusion and \eyeriss.}
	\label{fig:per-layer-eyeriss-energybreakdown}
	\vspace{-1ex}
\end{figure}
To understand the sources of the energy reduction, we break down the energy consumptions for each hardware component (compute units, on-chip SRAM buffers, register file, and off-chip DRAM memory).
Figure~\ref{fig:per-layer-eyeriss-energybreakdown} shows the per-component energy dissipation for \bitfusion and \eyeriss.
This figure should be considered with the energy reduction results from Figure~\ref{fig:improvement-over-eyeriss}.
Both accelerators consume more than \xx{80\%} of energy for on-chip and off-chip memory accesses.
The bit-level flexibility for memory accesses in \bitfusion significantly reduces energy consumption for both on-chip buffers (IBUF, OBUF, and WBUF in Figure~\ref{fig:arch}) and off-chip DRAM.
Furthermore, with bit-level flexibility, our buffers can hold more data at lower-bitwidths, effectively giving \bitfusion more on-chip storage capacity, which leads to fewer off-chip memory accesses.
\eyeriss employs local register files within each PE, which constitutes a significant portion of the energy consumption.
\bitfusion's systolic architecture avoids the need for register files and enforces explicit data sharing for inputs and partial results, as shown in Figure~\ref{fig:arch}.
Therefore, \bitfusion saves on Register File energy, but requires more SRAM accesses.
The combined effect of bit-level flexibility and the systolic organization of \bitbricks in the \bitfusion architecture provides an average energy savings of \eyerissEnergyAvg.
Off-chip DRAM accesses, however, are still a significant portion of \bitfusion's energy consumption and its share grows due to the significant reduction of compute and on-chip storage energy.

\subsubsection{Sensitivity Study} 
\niparagraph{\\Sensitivity to memory bandwidth.}
\begin{figure}
	\centering
	\includegraphics[width=1.0\linewidth]{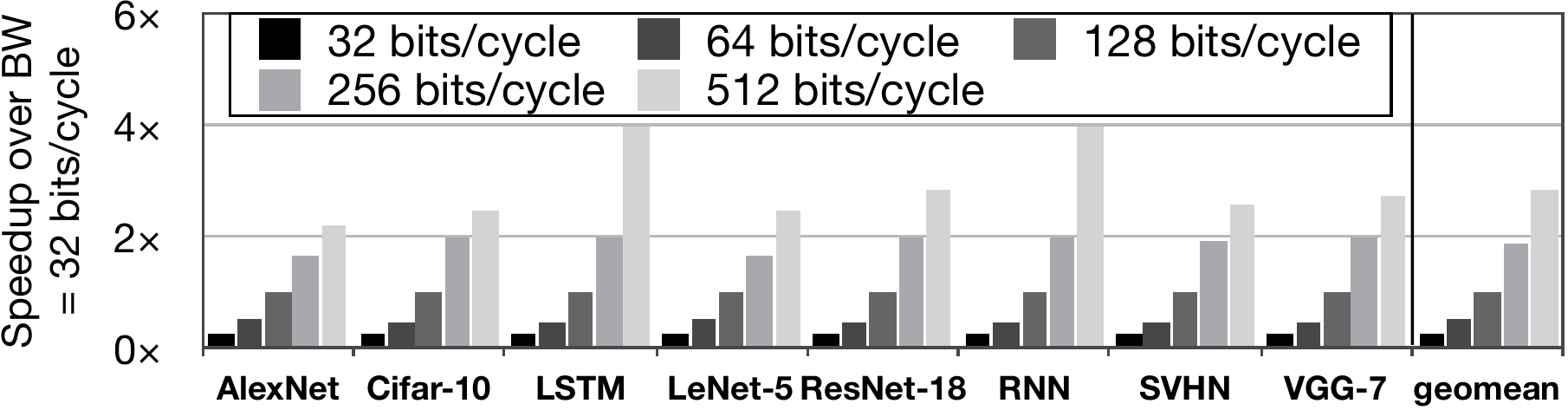}
	\vspace{-3.5ex}
	\caption{\bitfusion performance as the bandwidth changes.}
	\vspace{-3.5ex}
	\label{fig:bandwidth-sweep}
\end{figure}
Depending on the DNN topology, the impact of off-chip bandwidth on performance varies.
To understand the correlation between bandwidth and performance, we perform a sensitivity study for bandwidth.
Figure~\ref{fig:bandwidth-sweep} shows the performance improvements with \bitfusion as we change the bandwidth from 0.25$\times$ to 4$\times$ of the default value.
The baseline in this study the \bitfusion with the default bandwidth of \xx{128} bits per cycle.
On average, when we scale the bandwidth up to 4$\times$, \bitfusion provides \bench{1.6$\times$} speedup compared to the default setting, while with 0.25$\times$ bandwidth, the performance degrades \bench{60\%}.
Since CNN benchmarks see more opportunities for data reuse, they have less sensitivity to the bandwidth compared to the RNN benchmarks.
The two RNN benchmarks, \bench{LSTM} and \bench{RNN}, provide almost linearly-scaling speedup as they are bottlenecked by the bandwidth.

\begin{figure}
	\centering
	\includegraphics[width=1.0\linewidth]{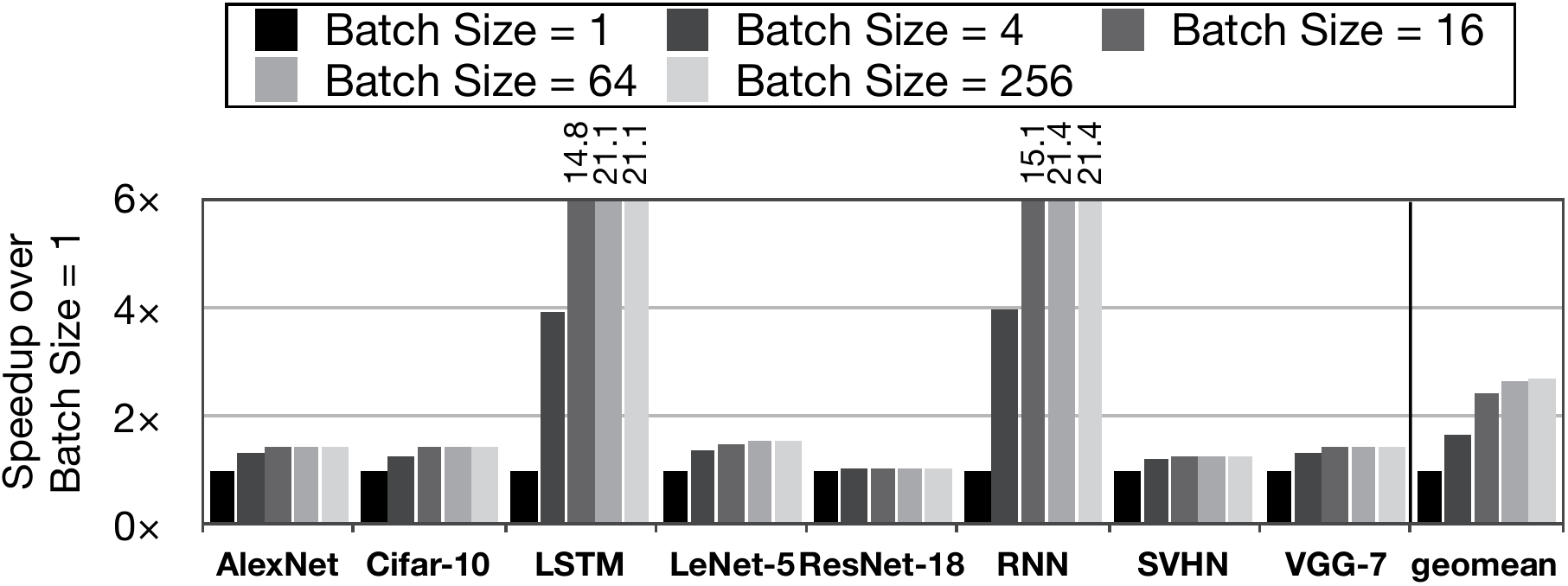}
	\vspace{-3.5ex}
	\caption{\bitfusion performance as the batch size increases.}
	\label{fig:batch-sweep}
	\vspace{-1ex}
\end{figure}

\niparagraph{Sensitivity to batch size.}
Batching amortizes the cost of weight reads by sharing weights across a batch of inputs.
Figure~\ref{fig:batch-sweep} shows how performance changes as we increase batch size from \xx{1} through \xx{256} with the batch size \xx{1} as the baseline (no batching).
Our default batch size is \xx{16}.
On average, \bitfusion with the batch size of \xx{256} engenders \xx{2.7$\times$} speedup with the highest speedup of \xx{21.4$\times$} from \bench{RNN}.
Since batching is effective when the bandwidth is limited and the performance is bandwidth-bound, the trends are similar to the bandwidth sensitivity results presented in Figure~\ref{fig:bandwidth-sweep}.
However, there is a marginal gain across all the benchmarks when the batch is increased from \xx{64} to \xx{256}, since beyond a batch size of \xx{64}, the bandwidth is sufficient to keep all the \fusionunits occupied.

\subsubsection{Comparison to GPUs}
\niparagraph{\\Performance comparison to GPUs.}
\begin{figure}
	\centering
	\includegraphics[width=1.0\linewidth]{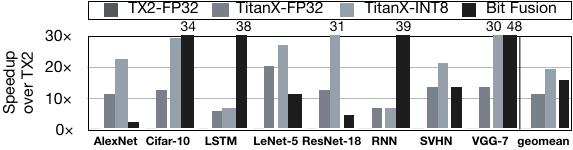}
	\vspace{-3ex}
	\caption{Performance comparison to GPUs.}
	\vspace{-3.5ex}
	\label{fig:gpu-performance-comparision}
\end{figure}
GPUs are the most widely-used general-purpose processors for DNNs.
We compare the performance of \bitfusion accelerators with two GPUs: (1) Tegra X2 (\bench{TX2}), and (2) Titan X based on the Pascal architecture (\titanxp), details of which are presented in Table~\ref{tab:platforms}.
As mentioned in the methodology section~\ref{sec:methodology}, we scale \bitfusion to match the 16~nm technology node of the GPUs, and use a total of 4096 \fusionunits.
Figure~\ref{fig:gpu-performance-comparision} shows the speedup of \bench{TitanX} and \bitfusion using the \bench{TX2} as the baseline.
\bench{TX2} does not support 8-bit mode natively.
Due to this lack of support, empirical results show slow down when the 8-bit instruction are used in \bench{TX2}.
As Figure~\ref{fig:gpu-performance-comparision} depicts, \bench{TitanX} in single-precision floating point (\xx{FP32}),  is, on average, \titanXpFPPerfAvgOverTX faster than \bench{TX2}.
The speedup grows to \titanXpPerfAvgOverTX when 8-bit mode is used.
While GPUs can benefit from using as low as 8-bits, \bitfusion can extract performance benefits for as low as 2-bit operations.
Using bit-level composability, \bitfusion provides a \bitfusionPerfAvgOverTX speedup over TX2.
The \bench{VGG-7} benchmark sees the maximum gains of \xx{30$\times$} and \xx{48$\times$} performance from \titanxp and \bitfusion, respectively.
The high degrees of parallelism in \bench{VGG-7} enables both \titanxp and \bitfusion to utilize all the available on-chip compute resources.
\bitfusion, while consuming 895 milliwatts of power, is only \xx{16$\%$}  slower than the 250-Watt \titanxp that uses $8$-bit computations, almost matching its performance.

\subsubsection{Comparison to \stripes}
\label{subsubsec:stipes-comparison}

\begin{figure}
	\centering
	\includegraphics[width=1.0\linewidth]{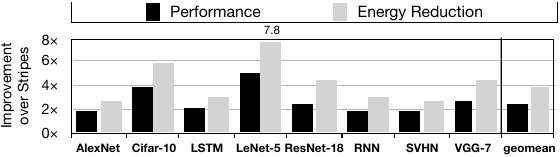}
	\vspace{-3ex}
	\caption{\bitfusion performance and energy improvements over \stripes.}
	\vspace{-3.5ex}
	\label{fig:improvement-over-stripes}
\end{figure}

\niparagraph{\\Performance compared to Stripes.}
Figure~\ref{fig:improvement-over-stripes} presents the performance and energy benefits of \bitfusion in comparison with \stripes.
On average, \bitfusion provides \stripesPerfAvg speedup over \stripes.
\stripes uses bit-serial computations to support variable bitwidths just for DNN weights.
As opposed to \stripes, the \bitfusion architecture offers dynamically composable \bitbricks to support flexible bitwidths for both inputs and weights in DNNs.
\bitfusion achieves the highest speedup of \xx{5.2$\times$} and lowest speedup of \xx{1.8$\times$} over \stripes for benchmarks \bench{LeNet-5} and \bench{AlexNet}, respectively.
\bench{ResNet-18} which is the most recent and the biggest of the benchmarks sees \xx{2.6$\times$} performance benefits as it can use low bitwidth on both operands.
\bench{AlexNet} uses 8-bit inputs/weights for the first convolution layer and the last fully-connected layer.
The two 8-bit layers limit the benefits of \bitfusion over \stripes.
Benchmark \bench{LeNet-5}, on the other hand, uses low bitwidths for both inputs and weights, resulting in the highest performance benefits with \bitfusion.

\niparagraph{Energy reduction compared to Stripes.}
Figure~\ref{fig:improvement-over-stripes} also depicts the improvement in energy when \bitfusion is compared to \stripes.
As mentioned, \bitfusion benefits from reduction in both computation and memory access at lower bitwidths for \emph{both} inputs and weights.
On average, \bitfusion reduces energy consumption by \stripesEnergyAvg over \stripes.
\bench{LeNet-5} sees the highest energy reduction of \xx{7.8$\times$}, while benchmark \bench{AlexNet} sees the least energy reduction of \xx{2.7$\times$} over \stripes.
For \bench{ResNet-18}, the energy is reduced by a factor of \xx{4$\times$}.

\bitfusion offers a fundamentally different approach from \stripes and explores the dimension of bit-level dynamic composabililty, which significantly improves performance and energy.

\vspace{-1ex}
\section{Related Work}
\label{sec:related}

A growing body of related works develop DNN accelerators.
\bitfusion fundamentally differs from prior work as it introduces and explores a new dimension of bit-level composable architectures that can dynamically match the bitwidth required by DNN operations.
\bitfusion aims to minimize both computations and communications in the finest granularity possible without compromising on the DNN accuracy.
Below, we discuss the most related work.

\niparagraph{Precision flexibility in DNNs.}
\stripes~\cite{stripes:micro:2016} and Tartan~\cite{tartan:arxiv:2017} use bit-serial compute units to provide precision flexibility for inputs at the cost of additional area overhead.
Both works provide performance and efficiency benefits that are proportional to the precision reduction for inputs.
We directly compare the benefits of \bitfusion to \stripes in Section~\ref{sec:eval}.
%
UNPU~\cite{unpu:isscc:2018} fabricates a bit-serial DNN accelerator at 65~nm, similar to \stripes~\cite{stripes:micro:2016}.
%
\loom~\cite{loom:arxiv:2017} uses bit-serial computation for precision flexibility.
\deeprecon~\cite{deeprecon:IJCNN:2017} skips stages of a fully-pipelined floating-point-multiplier to perform either one 16-bit, two 12-bit, or four 8-bit multiplications.
In contrast, the \fusionunits are spatial designs that use combinational logic to dynamically compose and decompose 2-bit multipliers (\bricks) to construct variable bitwidth multiply-add units.
Moons et al. propose aggressive voltage scaling techniques at low precision for increased energy efficiency at constant throughput by turning off parts of the multiplier~\cite{envision:isscc:2017, moons:vlsi:2016}.
As such, they do not offer fusion capabilities.
TPU~\cite{tpu:isca:2017} proposes a systolic architecture for DNNs and supports 8-bit and 16-bit precision.
This work, on the other hand, proposes an architecture that dynamically composes low-bitwidth compute units (\bricks) to match the bitwidth requirements of DNN layers.

\niparagraph{Binary DNN accelerators.}
Several inspiring works have explored ASIC and FPGA accelerators optimized for Binary DNNs.
FINN~\cite{finn:fpga:2017} uses FPGAs for accelerating Binary DNNs, while YodaNN~\cite{yodann:arxiv:2017} and BRein~\cite{brein:isscc:2017} propose an ASIC accelerator for binary DNNs.
Kim, et al.~\cite{binarydecompose:dac:2017} decompose the convolution weights for binary CNNs to improve performance and energy efficiency.
The above works focus solely on binary DNNs to achieve high performance at the cost of classification accuracy.
\bitfusion, on the other hand, flexibly matches the bitwidths of DNN operations for performance/energy benefits without losing accuracy.

\niparagraph{Sparse Accelerators for DNNs.}
EIE~\cite{eie:isca:2016}, Cambricon-X~\cite{cambricon-x:micro:2016}, Cnvlutin~\cite{cnvlutin:isca:2016}, and SCNN~\cite{scnn:isca:2017} explore the sparsity in the DNN layers and use zero-skipping to provide performance and energy-efficiency benefits.
Orthogonal to the works above, \bitfusion explores the dimension of bit-flexible accelerators for DNNs.

\niparagraph{Other ASIC accelerators for DNNs.}
DaDianNao~\cite{dadiannao:micro:2014} uses eDRAM to eliminate off-chip accesses and provide high performance and efficiency for DNNs.
PuDianNao~\cite{pudiannao:asplos:2015} is an accelerator designed for machine learning, but does not support CNNs.
Minerva~\cite{minerva:isca:2016} proposes operation pruning and data quantization techniques to reduce power consumption for ASIC acceleration.
\eyeriss~\cite{eyeriss:isca:2016, eyeriss:jssc:2017} presents an optimized row-stationary dataflow for DNNs to improve efficiency.
Tetris~\cite{tetris:asplos:2017} and Neurocube~\cite{neurocube:isca:2016} propose 3-D stacked DNN accelerators to provide high bandwidth for DNN operations.
ISAAC~\cite{isaac:isca:2016}, PipeLayer~\cite{pipelayer:hpca:2017}, and Prime~\cite{prime:isca:2016} use resistive RAM (ReRAM) for accelerating DNNs.
Ganax~\cite{ganax:isca:2018} uses a SIMD-MIMD architecture to support DNNs and generative models.
Snapea~\cite{snapea:isca:2018} employs early termination to skip computations.

\niparagraph{Instruction Sets for DNNs.}
Cambricon~\cite{cambricon:isca:2016} provides an ISA to express the different computations in a DNN using vector and matrix operations without significant loss in efficiency over DaDianNao.
DnnWeaver~\cite{dnnweaver:micro:2016} proposes a coarse grained ISA to express layers of DNNs, which are first translated to micro-codes for FPGA acceleration.
Unlike prior work, the \fusionisa proposed in the work is designed to enable bit-level flexibility for accelerating DNNs.
Further, the \fusionisa uses \code{loop} instructions with iterative semantics to significantly reduce instruction footprint.

\niparagraph{Code optimization techniques.}
Alwani, et. al~\cite{fused-cnn} propose layer-fusion, that combines multiple convolutional layers to save off-chip accesses for FPGA acceleration of CNNs.
Escher~\cite{escher:fccm:2017} proposes a CNN FPGA accelerator using flexible buffering that balances the off-chip accesses for inputs and weights in CNNs.
The above works have inspired the code-optimizations explored in this paper, however, the key contribution of this work is a bit-level flexible DNN accelerator.

\niparagraph{Software techniques for Binary/XNOR DNNs.}
QNN~\cite{qnn:arxiv:2016} shows that efficient GPU kernels for XNOR-based binary DNNs can provide up to 3.4$\times$ improvement in performance.
XNOR-Net~\cite{xnornet:arxiv:2017} shows that specialized libraries for Binary/XNOR-nets can achieve 58$\times$ performance on CPUs.
In contrast, \bitfusion is an ASIC accelerator architecture that supports a wide range of bitwidths (binary to 16-bits) for DNNs with no accuracy loss.

\niparagraph{Core Fusion and CLPs.} Core Fusion~\cite{corefusion} and CLPs~\cite{tflex} are dynamically configurable chip multiprocessors that a group of independent processors can fuse and form a more capable CPU.
In contrast to these inspiring works, \bitfusion performs the composition in the bit level rather than at the level of full-fledged cores.

\vspace{-1ex}
\section{Conclusion}
\label{sec:conclusion}

Deep neural networks use abundant computation, but can withstand very low bitwidth  operations without any loss in accuracy.
Leveraging this property of DNNs, we develop \bitfusion, a bit-level dynamically composable architecture, for their efficient acceleration.
The architecture comes with an ISA that enables the software to utilize this bit-level fusion capability to maximize the parallelism in computations and minimize the data transfer in the finest granularity possible.
We evaluate the benefits of \bitfusion by synthesizing the Verilog implementation of the proposed microarchitecture in 45~nm technology node and using cycle accurate simulations with eight real-world DNNs that require different bitwidths in their layers.
\bitfusion achieves significant speedup and energy benefits compared to state-of-the-art accelerators.
\vspace{-1.5ex}
\section{Acknowledgments}
We thank Amir Yazdanbaksh, Divya Mahajan, Jacob Sacks, and Payal Preet Bagga for insightful discussions and comments.
This work was in part supported by NSF awards CNS\#1703812, ECCS\#1609823, Air Force Office of Scientific Research (AFOSR) Young Investigator Program (YIP) award \#FA9550-17-1-0274, and gifts from Google, Microsoft, Xilinx, and Qualcomm.
%

\vspace{-1.5ex}
\bibliographystyle{ieeetr}
\bibliography{paper.bib}

\begin{thebibliography}{10}

\bibitem{eyeriss:isca:2016}
Y.-H. Chen, J.~Emer, and V.~Sze, ``Eyeriss: A spatial architecture for
  energy-efficient dataflow for convolutional neural networks,'' in {\em ISCA},
  2016.

\bibitem{stripes:micro:2016}
P.~Judd, J.~Albericio, T.~Hetherington, T.~M. Aamodt, and A.~Moshovos,
  ``Stripes: Bit-serial deep neural network computing,'' in {\em MICRO}, 2016.

\bibitem{eyeriss:jssc:2017}
Y.-H. Chen, T.~Krishna, J.~S. Emer, and V.~Sze, ``Eyeriss: An energy-efficient
  reconfigurable accelerator for deep convolutional neural networks,'' {\em
  JSSC}, 2017.

\bibitem{tetris:asplos:2017}
M.~Gao, J.~Pu, X.~Yang, M.~Horowitz, and C.~Kozyrakis, ``Tetris: Scalable and
  efficient neural network acceleration with 3d memory,'' in {\em ASPLOS},
  2017.

\bibitem{eie:isca:2016}
S.~Han, X.~Liu, H.~Mao, J.~Pu, A.~Pedram, M.~A. Horowitz, and W.~J. Dally,
  ``Eie: efficient inference engine on compressed deep neural network,'' in
  {\em ISCA}, 2016.

\bibitem{tartan:arxiv:2017}
A.~Delmas, S.~Sharify, P.~Judd, and A.~Moshovos, ``Tartan: Accelerating
  fully-connected and convolutional layers in deep learning networks by
  exploiting numerical precision variability,'' {\em arXiv}, 2017.

\bibitem{dadiannao:micro:2014}
Y.~Chen, T.~Luo, S.~Liu, S.~Zhang, L.~He, J.~Wang, L.~Li, T.~Chen, Z.~Xu,
  N.~Sun, {\em et~al.}, ``Dadiannao: A machine-learning supercomputer,'' in
  {\em MICRO}, 2014.

\bibitem{diannao:asplos:2014}
T.~Chen, Z.~Du, N.~Sun, J.~Wang, C.~Wu, Y.~Chen, and O.~Temam, ``Diannao: a
  small-footprint high-throughput accelerator for ubiquitous
  machine-learning,'' in {\em ASPLOS}, 2014.

\bibitem{pudiannao:asplos:2015}
D.~Liu, T.~Chen, S.~Liu, J.~Zhou, S.~Zhou, O.~Teman, X.~Feng, X.~Zhou, and
  Y.~Chen, ``Pudiannao: A polyvalent machine learning accelerator,'' in {\em
  ASPLOS}, 2015.

\bibitem{shidiannao:isca:2015}
Z.~Du, R.~Fasthuber, T.~Chen, P.~Ienne, L.~Li, T.~Luo, X.~Feng, Y.~Chen, and
  O.~Temam, ``Shidiannao: shifting vision processing closer to the sensor,'' in
  {\em ISCA}, 2015.

\bibitem{neurocube:isca:2016}
D.~Kim, J.~Kung, S.~Chai, S.~Yalamanchili, and S.~Mukhopadhyay, ``Neurocube: A
  programmable digital neuromorphic architecture with high-density 3d memory,''
  in {\em ISCA}, 2016.

\bibitem{minerva:isca:2016}
B.~Reagen, P.~Whatmough, R.~Adolf, S.~Rama, H.~Lee, S.~K. Lee, J.~M.
  Hern{\'a}ndez-Lobato, G.-Y. Wei, and D.~Brooks, ``Minerva: Enabling
  low-power, highly-accurate deep neural network accelerators,'' in {\em ISCA},
  2016.

\bibitem{cnvlutin:isca:2016}
J.~Albericio, P.~Judd, T.~Hetherington, T.~Aamodt, N.~E. Jerger, and
  A.~Moshovos, ``Cnvlutin: ineffectual-neuron-free deep neural network
  computing,'' in {\em ISCA}, 2016.

\bibitem{cambricon:isca:2016}
S.~Liu, Z.~Du, J.~Tao, D.~Han, T.~Luo, Y.~Xie, Y.~Chen, and T.~Chen,
  ``Cambricon: An instruction set architecture for neural networks,'' in {\em
  ISCA}, 2016.

\bibitem{cambricon-x:micro:2016}
S.~Zhang, Z.~Du, L.~Zhang, H.~Lan, S.~Liu, L.~Li, Q.~Guo, T.~Chen, and Y.~Chen,
  ``Cambricon-x: An accelerator for sparse neural networks,'' in {\em MICRO},
  2016.

\bibitem{240gopsmobile:cvprw:2014}
V.~Gokhale, J.~Jin, A.~Dundar, B.~Martini, and E.~Culurciello, ``A 240 g-ops/s
  mobile coprocessor for deep neural networks,'' in {\em CVPRW}, 2014.

\bibitem{iotcnn:isscc:2016}
J.~Sim, J.~S. Park, M.~Kim, D.~Bae, Y.~Choi, and L.~S. Kim, ``14.6 a 1.42tops/w
  deep convolutional neural network recognition processor for intelligent ioe
  systems,'' in {\em ISSCC}, 2016.

\bibitem{riscintconv:date:2015}
F.~Conti and L.~Benini, ``A ultra-low-energy convolution engine for fast
  brain-inspired vision in multicore clusters,'' in {\em DATE}, 2015.

\bibitem{deepburning:dac:2016}
Y.~Wang, J.~Xu, Y.~Han, H.~Li, and X.~Li, ``Deepburning: Automatic generation
  of fpga-based learning accelerators for the neural network family,'' in {\em
  DAC}, 2016.

\bibitem{cbrain:dac:2016}
L.~Song, Y.~Wang, Y.~Han, X.~Zhao, B.~Liu, and X.~Li, ``C-brain: A deep
  learning accelerator that tames the diversity of cnns through adaptive
  data-level parallelization,'' in {\em DAC}, 2016.

\bibitem{dnnoptimizing:fpga:2015}
C.~Zhang, P.~Li, G.~Sun, Y.~Guan, B.~Xiao, and J.~Cong, ``Optimizing fpga-based
  accelerator design for deep convolutional neural networks,'' in {\em FPGA},
  2015.

\bibitem{dnnweaver:micro:2016}
H.~Sharma, J.~Park, D.~Mahajan, E.~Amaro, J.~K. Kim, C.~Shao, A.~Misra, and
  H.~Esmaeilzadeh, ``From high-level deep neural models to fpgas,'' in {\em
  MICRO}, 2016.

\bibitem{fusedlayercnn:micro:2016}
M.~Alwani, H.~Chen, M.~Ferdman, and P.~Milder, ``Fused-layer cnn
  accelerators,'' in {\em MICRO}, 2016.

\bibitem{openclcnn:fpga:2016}
N.~Suda, V.~Chandra, G.~Dasika, A.~Mohanty, Y.~Ma, S.~Vrudhula, J.-s. Seo, and
  Y.~Cao, ``Throughput-optimized opencl-based fpga accelerator for large-scale
  convolutional neural networks,'' in {\em FPGA}, 2016.

\bibitem{embedfpgacnn:fpga:2016}
J.~Qiu, J.~Wang, S.~Yao, K.~Guo, B.~Li, E.~Zhou, J.~Yu, T.~Tang, N.~Xu,
  S.~Song, {\em et~al.}, ``Going deeper with embedded fpga platform for
  convolutional neural network,'' in {\em FPGA}, 2016.

\bibitem{isaac:isca:2016}
A.~Shafiee, A.~Nag, N.~Muralimanohar, R.~Balasubramonian, J.~P. Strachan,
  M.~Hu, R.~S. Williams, and V.~Srikumar, ``Isaac: A convolutional neural
  network accelerator with in-situ analog arithmetic in crossbars,'' in {\em
  ISCA}, 2016.

\bibitem{prime:isca:2016}
P.~Chi, S.~Li, C.~Xu, T.~Zhang, J.~Zhao, Y.~Liu, Y.~Wang, and Y.~Xie, ``Prime:
  A novel processing-in-memory architecture for neural network computation in
  reram-based main memory,'' in {\em ISCA}, 2016.

\bibitem{pipelayer:hpca:2017}
L.~Song, X.~Qian, H.~Li, and Y.~Chen, ``Pipelayer: A pipelined reram-based
  accelerator for deep learning,'' in {\em HPCA}, 2017.

\bibitem{brainwave:hotchips:2017}
E.~Chung, J.~Fowers, K.~Ovtcharov, M.~Papamichael, A.~Caulfield, T.~Massengil,
  M.~Liu, D.~Lo, S.~Alkalay, M.~Haselman, C.~Boehn, O.~Firestein, A.~Forin,
  K.~S. Gatlin, M.~Ghandi, S.~Heil, K.~Holohan, T.~Juhasz, R.~K. Kovvuri,
  S.~Lanka, F.~van Megen, D.~Mukhortov, P.~Patel, S.~Reinhardt, A.~Sapek,
  R.~Seera, B.~Sridharan, L.~Woods, P.~Yi-Xiao, R.~Zhao, and D.~Burger,
  ``Accelerating persistent neural networks at datacenter scale,'' in {\em
  HotChips}, 2017.

\bibitem{tpu:isca:2017}
N.~P. Jouppi, C.~Young, N.~Patil, D.~Patterson, G.~Agrawal, R.~Bajwa, S.~Bates,
  S.~Bhatia, N.~Boden, A.~Borchers, {\em et~al.}, ``In-datacenter performance
  analysis of a tensor processing unit,'' in {\em ISCA}, 2017.

\bibitem{apple-a11bionic:wiki:2017}
``Apple a11-bionic.'' \url{https://en.wikipedia.org/wiki/Apple_A11}.

\bibitem{dorefa:arxiv:2016}
S.~Zhou, Z.~Ni, X.~Zhou, H.~Wen, Y.~Wu, and Y.~Zou, ``Dorefa-net: Training low
  bitwidth convolutional neural networks with low bitwidth gradients,'' {\em
  arXiv}, 2016.

\bibitem{zhu2016trained}
C.~Zhu, S.~Han, H.~Mao, and W.~J. Dally, ``Trained ternary quantization,'' {\em
  arXiv}, 2016.

\bibitem{li2016ternary}
F.~Li, B.~Zhang, and B.~Liu, ``Ternary weight networks,'' {\em arXiv}, 2016.

\bibitem{qnn:arxiv:2016}
I.~Hubara, M.~Courbariaux, D.~Soudry, R.~El-Yaniv, and Y.~Bengio, ``Quantized
  neural networks: Training neural networks with low precision weights and
  activations,'' {\em arXiv}, 2016.

\bibitem{wrpn}
A.~K. Mishra, E.~Nurvitadhi, J.~J. Cook, and D.~Marr, ``{WRPN:} wide
  reduced-precision networks,'' {\em arXiv}, 2017.

\bibitem{envision:isscc:2017}
B.~Moons, R.~Uytterhoeven, W.~Dehaene, and M.~Verhelst, ``Dvafs: Trading
  computational accuracy for energy through
  dynamic-voltage-accuracy-frequency-scaling,'' in {\em DATE}, 2017.

\bibitem{loom:arxiv:2017}
S.~Sharify, A.~D. Lascorz, P.~Judd, and A.~Moshovos, ``Loom: Exploiting weight
  and activation precisions to accelerate convolutional neural networks,'' {\em
  arXiv}, 2017.

\bibitem{unpu:isscc:2018}
J.~Lee, C.~Kim, S.~Kang, D.~Shin, S.~Kim, and H.-J. Yoo, ``Unpu: A 50.6 tops/w
  unified deep neural network accelerator with 1b-to-16b fully-variable weight
  bit-precision,'' in {\em ISSCC}, 2018.

\bibitem{alexnet}
A.~Krizhevsky, ``One weird trick for parallelizing convolutional neural
  networks,'' {\em arXiv}, 2014.

\bibitem{svhn:nips:2011}
Y.~Netzer, T.~Wang, A.~Coates, A.~Bissacco, B.~Wu, and A.~Y. Ng, ``Reading
  digits in natural images with unsupervised feature learning,'' in {\em NIPS
  workshop on deep learning and unsupervised feature learning}, 2011.

\bibitem{cifar10}
A.~Krizhevsky and G.~Hinton, ``Learning multiple layers of features from tiny
  images,'' {\em Computer Science Department, University of Toronto, Tech.
  Rep}, 2009.

\bibitem{lenet-5}
Y.~LeCun, L.~Bottou, Y.~Bengio, and P.~Haffner, ``Gradient-based learning
  applied to document recognition,'' {\em Proceedings of the IEEE}, vol.~86,
  no.~11, pp.~2278--2324, 1998.

\bibitem{vgg}
K.~Simonyan and A.~Zisserman, ``Very deep convolutional networks for
  large-scale image recognition,'' {\em arXiv}, 2014.

\bibitem{resnet}
K.~He, X.~Zhang, S.~Ren, and J.~Sun, ``Deep residual learning for image
  recognition,'' in {\em CVPR}, 2016.

\bibitem{lstm}
S.~Hochreiter and J.~Schmidhuber, ``Long short-term memory,'' {\em Neural
  computation}, 1997.

\bibitem{penn-treebank}
M.~P. Marcus, M.~A. Marcinkiewicz, and B.~Santorini, ``Building a large
  annotated corpus of english: The penn treebank,'' {\em Computational
  linguistics}, 1993.

\bibitem{cactip}
S.~Li, K.~Chen, J.~H. Ahn, J.~B. Brockman, and N.~P. Jouppi, ``{CACTI-P:
  Architecture-level Modeling for SRAM-based Structures with Advanced Leakage
  Reduction Techniques},'' in {\em ICCAD}, 2011.

\bibitem{tensorrt}
``Nvidia tensor rt 4.0.'' \url{https://developer.nvidia.com/tensorrt}.

\bibitem{dark_silicon:isca}
H.~Esmaeilzadeh, E.~Blem, R.~St.~Amant, K.~Sankaralingam, and D.~Burger, ``Dark
  silicon and the end of multicore scaling,'' in {\em ISCA}, 2011.

\bibitem{deeprecon:IJCNN:2017}
T.~Rzayev, S.~Moradi, D.~H. Albonesi, and R.~Manohar, ``Deeprecon: Dynamically
  reconfigurable architecture for accelerating deep neural networks,'' {\em
  IJCNN}, 2017.

\bibitem{moons:vlsi:2016}
B.~Moons and M.~Verhelst, ``A 0.3--2.6 tops/w precision-scalable processor for
  real-time large-scale convnets,'' in {\em VLSI-Circuits}, 2016.

\bibitem{finn:fpga:2017}
Y.~Umuroglu, N.~J. Fraser, G.~Gambardella, M.~Blott, P.~Leong, M.~Jahre, and
  K.~Vissers, ``Finn: A framework for fast, scalable binarized neural network
  inference,'' in {\em FPGA}, 2017.

\bibitem{yodann:arxiv:2017}
R.~Andri, L.~Cavigelli, D.~Rossi, and L.~Benini, ``Yodann: An ultra-low power
  convolutional neural network accelerator based on binary weights,'' {\em
  arXiv}, 2016.

\bibitem{brein:isscc:2017}
K.~Ando, K.~Ueyoshi, K.~Orimo, H.~Yonekawa, S.~Sato, H.~Nakahara, M.~Ikebe,
  T.~Asai, S.~Takamaeda-Yamazaki, T.~Kuroda, {\em et~al.}, ``Brein memory: A
  13-layer 4.2 k neuron/0.8 m synapse binary/ternary reconfigurable in-memory
  deep neural network accelerator in 65 nm cmos,'' in {\em VLSI}, 2017.

\bibitem{binarydecompose:dac:2017}
H.~Kim, J.~Sim, Y.~Choi, and L.-S. Kim, ``A kernel decomposition architecture
  for binary-weight convolutional neural networks,'' in {\em DAC}, 2017.

\bibitem{scnn:isca:2017}
A.~Parashar, M.~Rhu, A.~Mukkara, A.~Puglielli, R.~Venkatesan, B.~Khailany,
  J.~Emer, S.~W. Keckler, and W.~J. Dally, ``{SCNN: An Accelerator for
  Compressed-sparse Convolutional Neural Networks},'' in {\em ISCA}, 2017.

\bibitem{ganax:isca:2018}
A.~Yazdanbakhsh, H.~Falahati, P.~J. Wolfe, K.~Samadi, H.~Esmaeilzadeh, and
  N.~S. Kim, ``{GANAX: A Unified SIMD-MIMD Acceleration for Generative
  Adversarial Network},'' in {\em ISCA}, 2018.

\bibitem{snapea:isca:2018}
V.~Aklaghi, A.~Yazdanbakhsh, K.~Samadi, H.~Esmaeilzadeh, and R.~K.~Gupta,
  ``Snapea: Predictive early activation for reducing computation in deep
  convolutional neural networks,'' in {\em ISCA}, 2018.

\bibitem{fused-cnn}
M.~Alwani, H.~Chen, M.~Ferdman, and P.~Milder, ``Fused-layer cnn accelerator,''
  in {\em MICRO}, 2016.

\bibitem{escher:fccm:2017}
Y.~Shen, M.~Ferdman, and P.~Milder, ``Escher: A cnn accelerator with flexible
  buffering to minimize off-chip transfer,'' in {\em FCCM}, 2017.

\bibitem{xnornet:arxiv:2017}
M.~Rastegari, V.~Ordonez, J.~Redmon, and A.~Farhadi, ``Xnor-net: Imagenet
  classification using binary convolutional neural networks,'' {\em arXiv},
  2016.

\bibitem{corefusion}
E.~Ipek, M.~Kirman, N.~Kirman, and J.~F. Martinez, ``Core fusion: accommodating
  software diversity in chip multiprocessors,'' in {\em ISCA}, 2007.

\bibitem{tflex}
C.~Kim, S.~Sethumadhavan, M.~Govindan, N.~Ranganathan, D.~Gulati, D.~Burger,
  and S.~W. Keckler, ``Composable lightweight processors,'' in {\em MICRO},
  2007.

\end{thebibliography}

\setlength{\paperheight}{11in}
\setlength{\paperwidth}{8.5in}

\end{document}